\documentclass[conference]{IEEEtran}
\IEEEoverridecommandlockouts
\usepackage{cite}
\usepackage{amsmath,amssymb,amsfonts}
\usepackage{algorithmic}
\usepackage{graphicx}
\usepackage{textcomp}
\usepackage{orcidlink}
\usepackage{soul} 
\usepackage{xcolor}
\usepackage[table,xcdraw]{xcolor}
\usepackage{multirow}
\usepackage{placeins}
\usepackage{nameref} 
\usepackage{fancyhdr}
\def\BibTeX{{\rm B\kern-.05em{\sc i\kern-.025em b}\kern-.08em
    T\kern-.1667em\lower.7ex\hbox{E}\kern-.125emX}}
\begin{document}

\title{XLinear: Frequency-Enhanced MLP with CrossFilter for Robust Long-Range Forecasting}

\author{\IEEEauthorblockN{1\textsuperscript{st} Xiang Ao}
\IEEEauthorblockA{\textit{School of Software Engineering} \\
\textit{Beijing Jiaotong University}\\
Beijing, China \\
ao.xiang.axel@outlook.com\orcidlink{https://orcid.org/0000-0002-8637-8375}}

}

\maketitle
\thispagestyle{fancy}

\begin{abstract}
Time series forecasters are widely used across various domains. Among them,  MLP(multi-layer perceptron)-based forecasters have been proven to be more robust to noise compared to Transformer-based forecasters. However, MLP struggles to capture complex features, resulting in limitations on capturing long-range dependencies. To address this challenge, we propose the XLinear, a MLP-based forecaster for long-range forecasting. Firstly, we decompose the time series into trend and seasonal components. For the trend component which contains long-range characteristics, we design EFA (Enhanced Frequency Attention) to capture long-term dependencies by leveraging frequency-domain operations. Additionally, a CrossFilter Block is proposed for the seasonal component to keep the model’s robustness to noise, avoiding the problems of low robustness caused by the attention mechanism. Experimental results demonstrate that XLinear achieves state-of-the-art performance on test datasets. While keeping the lightweight architecture and high robustness of MLP-based models, our forecaster outperforms other MLP-based forecasters in capturing long-range dependencies.
\end{abstract}


\section{Introduction}
Time series forecasting is a prominent area of research in statistics and machine learning. With the development of data science, time series forecasting algorithm has found widespread application in various domains such as forecasting meteorological factors for weather prediction, stock price prediction in finance, estimating time of arriving in traffic domain, forecasting clinical risk of influenza like COVID-19 et al.\cite{1,2,3,4} To improve the accuracy of time series forecasting, extensive research has been conducted on deep learning-based forecasters for time series forecasting, leading to the emergence of numerous forecasters based on deep learning dedicated to this task.

Previously, most of forecasters are based on Recurrent Neural Network(RNN) like LSTM.\cite{5}\cite{6} Recently, forecasters based on Convolution Neural Network(CNN) received particularly attention such as TCN ModernTCN.\cite{7}\cite{8} Since the proposal of the Transformer architecture, many studies have attempted to apply this excellent structure for time series forecasting such as Autoformer, PatchTST, iTransformer, et al.\cite{9,10,11,12} Additionally, methods based on multi-layer perceptrons (MLPs) achieved excellent performance with lower parameter counts. Especially, some excellent MLP-based forecasters outperformed the state-of-the-art Transformer-based forecasters such as DLinear and FilterNet.\cite{13}\cite{14} Since the research of DLinear proved that light and simple MLP-based forecasters could perform better than complex Transformer-based forecasters, researchers have tried to improve MLP-based forecasters.

However, MLP-based forecasters still suffer from capturing long-range dependencies and perform poorly when faced with a complex data set.\cite{14} To overcome the limitation of MLP-based forecasters, we proposed XLinear, which perform well in predict results in the far future while keeping the advantages of lightweightness and simplicity of MLP-based forecasters.

Forecasters based on MLP are constrained by the receptive field of linear connections, making it difficult to learn complex features.\cite{8} While Transformer-based forecasters can effectively capture complex features through attention mechanisms, their high computational resource consumption and vulnerability to high-frequency signals pose challenges when directly fused with MLPs.\cite{15} Therefore, we propose applying the attention mechanism only on local component of time series to avoid these limitation, enabling the attention mechanism to fully exploit its strengths. Because the trend component of time series represent long-term characteristics, we decompose time-series data into trend and seasonal components and employ the attention mechanism on the trend component.\cite{16}\cite{17}

To further improve the attention mechanism’s capability to capture trends of time series, we propose an EFA (Enhanced Frequency Attention) block. This block performs frequency-domain operations on queries (Q) and keys (K). Fast Fourier transform (FFT) is used for frequency-domain transformation, followed by element-wise multiplication. Leveraging the low-frequency (trend) and high-frequency (noise/short-term fluctuations) characteristics of time-series data, this approach preserves trends while suppressing noise. An exponential activation function is also applied to values (V) to further amplify trend signals. 

Despite the effectiveness of EFA, the mere introduction of the attention mechanism may incur robustness issues inherent to attention-based architectures. Although filter blocks have been widely adopted in MLP-based forecasters, conventional filters suffer from limitations in stability and generalizability across diverse datasets—from simple to complex.\cite{14}\cite{15} To address this, we propose a CrossFilter block, a filter block that suits to most of the datasets. To mitigate cross-noise interference during fusion, we employ frequency-domain multiplication combined with GELU (Gaussian Error Linear Unit) to integrate the two filters. The CrossFilter is applied to the seasonal component, which typically contains the majority of noise. Finally, the trend and seasonal components are recombined and fed into an MLP for predictive output.

Owing to the frequency-enhanced design, XLinear demonstrates superior performance over other MLP-based forecasters in capturing long-range dependencies. Moreover, the advanced filtering mechanism enables XLinear to exhibit remarkable robustness across both complex and simple datasets. In comparative experiments, XLinear achieved outstanding prediction accuracy, particularly in long-sequence forecasting tasks. Additionally, ablation experiments further validated the effectiveness of our filtering module and frequency enhancement mechanisms. To summarize, the main contributions of this work are as follows:

\begin{itemize}
\item We introduced the XLinear network, which use attention mechanism on the trend component of time-series to capture the long-range dependencies while keep lightweightness simplicity.
	
\item We proposed the EFA block based on frequency-domain operations and exponential activation to capture the trends of time-series and an improved filter namely CrossFilter block to improve the noise robustness and generality of XLinear.

\item We compare XLinear with state-of-the-art (SOTA) forecasters on eight datasets. Experimental results demonstrate that XLinear performed better than other MLP-based forecaster especially on long-range forecasting while keeping the advantage of MLP-based forecasters. 
\end{itemize}

\begin{figure*}[h]
    \centering
    \includegraphics[width=1\linewidth]{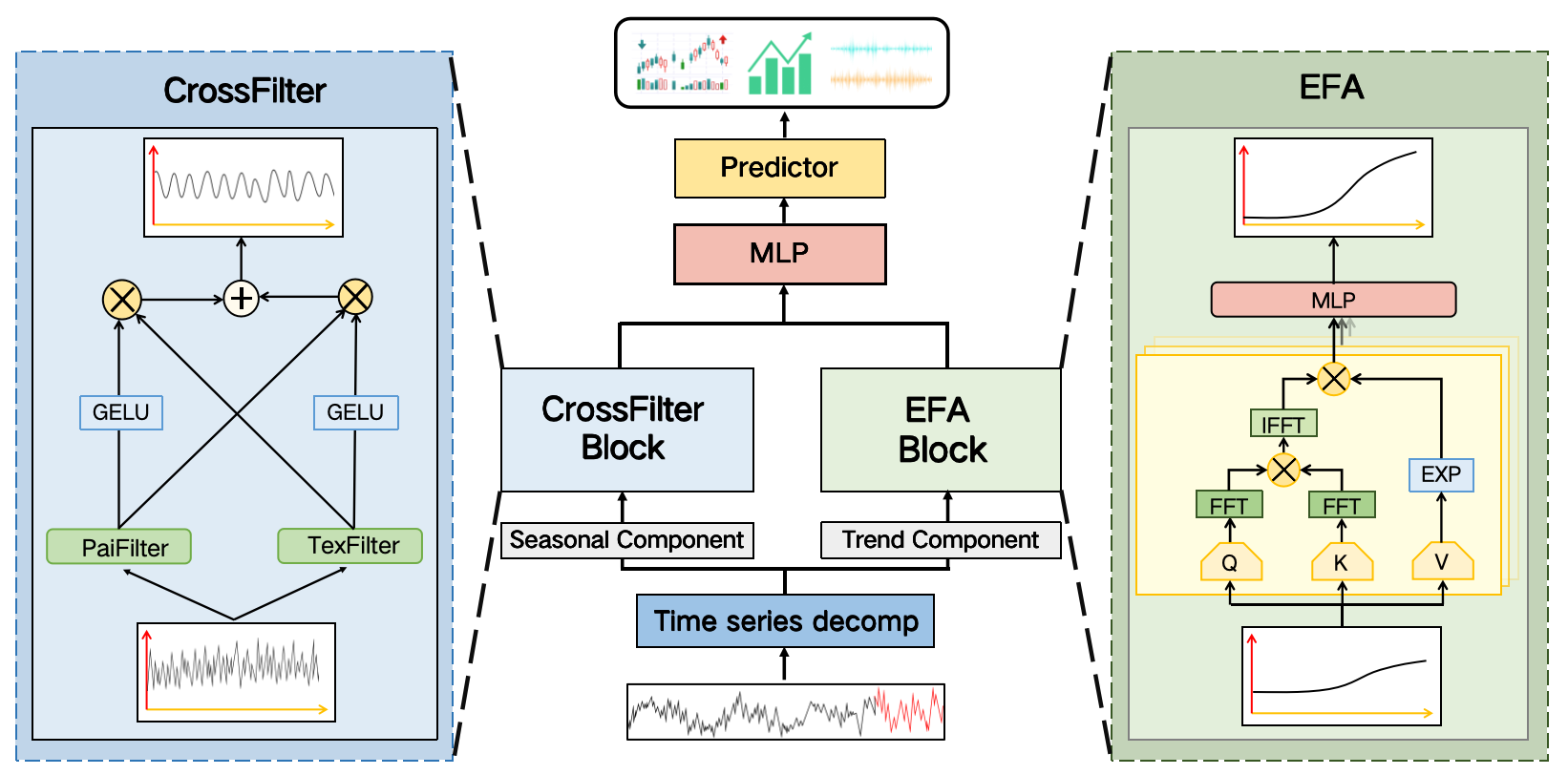}
    \caption{Framework of our XLinear. The 'Time series decomp' is moving average time series decomposition in TimesNet.\cite{17} The seasonal component is input to CrossFilter Block and trend component is input into EFA Block. The outputs of these two blocks are concatenated and then processed by an MLP layer.}
    \label{fig:XLinear}
\end{figure*}

\section{PRELIMINARIES}
\label{sec:pre}
\subsection{Time Series Decomposition}
Classically, time series decomposition separates data into trend and seasonal components.\cite{16} Mathematically, a time series \( x_t \) can be decomposed as:

\begin{equation}
    x_t = T_t + S_t + N_t
\end{equation}
where \( T_t \) represents the trend component, characterized as the low-frequency component with minimal short-term fluctuations, which inherently captures long-term dependencies. The seasonal component \( S_t \) comprises high-frequency periodic variations that encode short-term periodic dependencies. Notably, since noise \( N_t \) predominantly manifests as high-frequency signals, it is often retained within the seasonal component during decomposition.

In this decomposition approach, the trend component is extracted via a moving average operation. For a time series \( x \) with kernel size \( k \), the moving average \( T \) is computed as:

\begin{equation}
  T_t = \frac{1}{k} \sum_{i=0}^{k-1} x  
\end{equation}
This operation effectively smooths out short-term fluctuations to highlight the underlying trend. The residual component, containing both seasonal variations and noise, is then obtained by:

\begin{equation}
    R_t = x_t - T_t
\end{equation}
where \( R_t = S_t + N_t \) aggregates the high-frequency components of the original time series.

\subsection{Attention Mechanism}
The attention mechanism serves as the core component of the Transformer architecture.\cite{9} By leveraging a large receptive field, this mechanism effectively captures fine-grained data features, thereby enhancing the model's capacity to model complex interdependencies. Empirical studies have demonstrated that attention mechanisms excel at capturing long-range dependencies in sequential data.\cite{12} However, such mechanisms often encounter two prominent challenges.

\textbf{Parameter Overhead:}
In the widely adopted multi-head attention module, the parameter count can be formulated as:\begin{equation}
\text{P} = 3 \times h \times d_{\text{model}} \times d_k,
\end{equation}
where $P$ denotes the total number of parameters, $h$ represents the number of attention heads, \(d_{\text{model}}\) signifies the model's hidden dimension, and \(d_k\) corresponds to the dimension of each attention head. Although full-attention mechanisms excel at capturing complex information, they lack explicit frequency-domain enhancement objectives. To model long-range dependencies, full attention typically relies on large parameter spaces.

\textbf{Robustness Deficiency:}
Empirical studies have revealed that attention mechanisms are susceptible to noise interference.\cite{14}\cite{15} This inherent fragility necessitates the implementation of tailored strategies to enhance their resilience in real-world scenarios.

In XLinear, we tackle the aforementioned challenges by locally applying a specifically designed attention mechanism in tandem with filter operations.

\subsection{Fast Fourier Transform}
The Fast Fourier Transform (FFT) enables efficient computation of the Discrete Fourier Transform (DFT) through a divide-and-conquer strategy, providing core support for optimizing attention mechanisms. In models such as Transformers, the traditional scaled dot-product attention has a time complexity of \(O(N^2)\) (where \(N\) is the sequence length), resulting in significant efficiency bottlenecks when processing long sequences.

The divide-and-conquer strategy of FFT is based on the parity decomposition of sequences: For a sequence \(x[n]\) of length \(N\), it is first split into an odd-indexed subsequence \(x[1], x[3], \ldots, x[N-1]\) and an even-indexed subsequence \(x[0], x[2], \ldots, x[N-2]\). The DFTs of these two subsequences are computed separately (denoted as \(X_{\text{odd}}[k]\) and \(X_{\text{even}}[k]\)). Using the periodicity of the complex exponential function \(e^{-i2\pi(k+N/2)/N} = -e^{-i2\pi k/N}\), the DFT of the original sequence can be expressed as:

\begin{equation}
X[k] = X_{\text{even}}[k] + W_N^k X_{\text{odd}}[k]
\end{equation}
\begin{equation}
X[k+N/2] = X_{\text{even}}[k] - W_N^k X_{\text{odd}}[k]
\end{equation}
where \(W_N^k = e^{-i2\pi k/N}\) is the twiddle factor.This decomposition transforms a problem of size \(N\) into two subproblems of size \(N/2\). After recursively processing down to the smallest scale, the results are merged backward, ultimately reducing the time complexity of DFT from \(O(N^2)\) to \(O(N \log N)\). When applied to attention mechanisms, sequences \(x\) and weight matrices \(A\) are transformed into the Fourier domain using FFT (\(X = \text{FFT}(x)\), \(A' = \text{FFT}(A)\)). Leveraging the convolution theorem, dot product operations are converted into element-wise multiplications. After transforming back to the original domain via inverse FFT, the time complexity of attention computation is similarly reduced to \(O(N \log N)\), significantly improving the efficiency of long sequence processing.

\section{METHODOLOGY}
\subsection{Model Structure}
\sethlcolor{yellow}
We proposed XLinear to capture long-range dependencies while maintaining robustness. Figure \ref{fig:XLinear}. illustrates the architecture of XLinear. First, the time series data are decomposed into trend and seasonal components via moving average-based time series decomposition. This is a classic method, which has been mentioned in the section\ref{sec:pre} Then, the trend component, which encoding long-range dependencies, is fed into an EFA block. This attention block is used locally to avoid parameter overhead and typically designed for frequency enhancement. Conversely, the seasonal component is processed by a CrossFilter block to mitigate high-frequency noise, thereby enhancing model robustness. Finally, the two processed components are concatenated and passed through an MLP to generate predictions.

\subsection{EFA(Enhanced Frequency Attention) Block}
\begin{figure}
    \centering
    \includegraphics[width=1\linewidth]{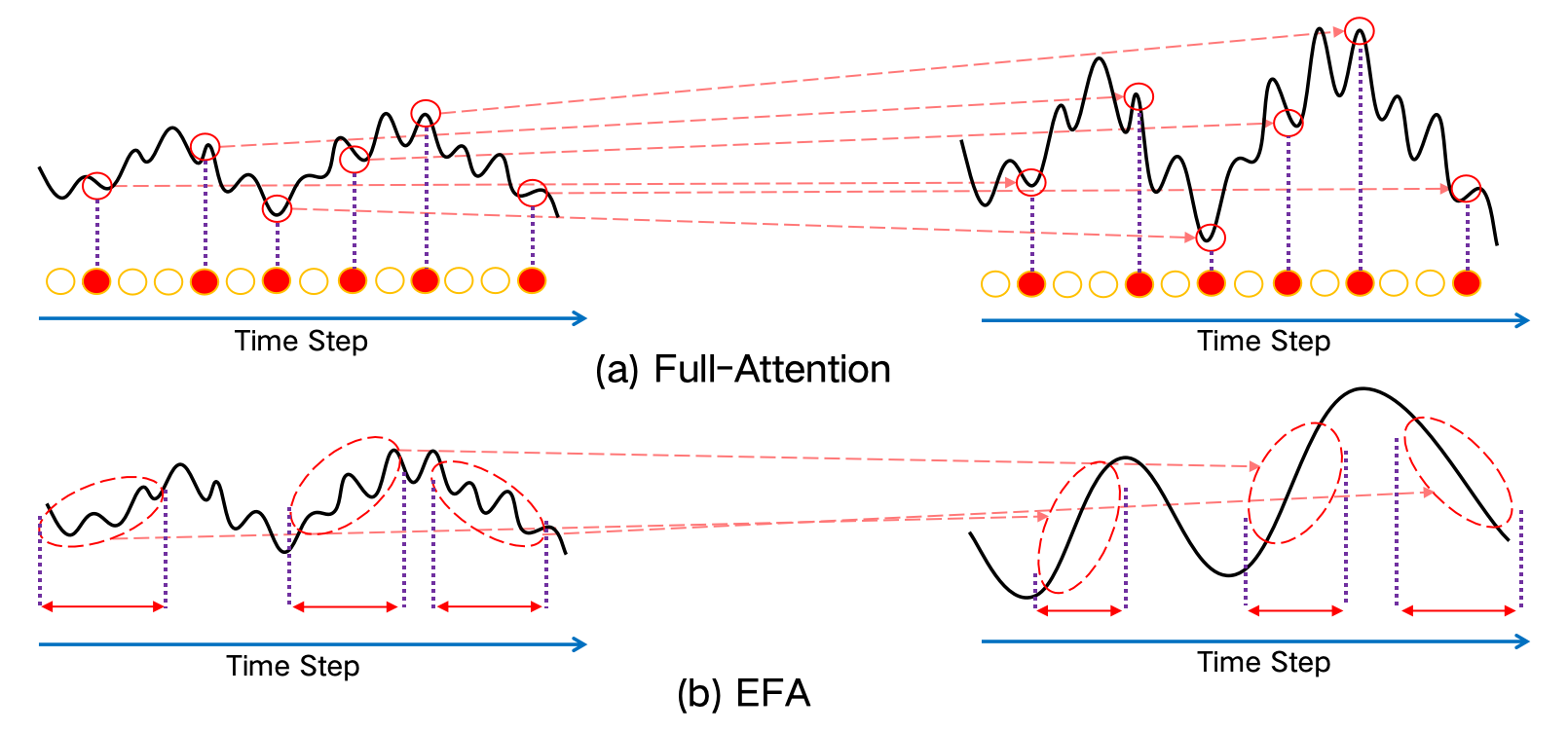}
    \caption{EFA vs. other attention. (a) Other attention:every components of the series are strengthened. (b) EFA: only low-frequency component is strengthened.}
    \label{fig:attention}
\end{figure}

EFA targets low-frequency information explicitly, focusing on enhancing the trend component of time series. The distinction between full attention and EFA is illustrated in Figure \ref{fig:attention}. To strengthen trend modeling, EFA employs frequency-aware weighting to selectively amplify low-frequency components. This is achieved through frequency-domain operations that attenuate high-frequency noise while reinforcing trend-related low-frequency signals. Additionally, a non-linear activation function is incorporated to further refine trend representation, enabling more robust capture of underlying temporal dynamics. 

In EFA, the input $X_{trend} \in \mathbb{R}^{L \times c}$ is the trend component of input series $X \in \mathbb{R}^{L \times c}$ after decomposition. Where $L$ is the input length and the $c$ is the channel counts. The $X_{trend}$ is encoded into Query ($Q$), Key ($K$), and Value ($V$) via linear projection: 
\begin{equation}
    Q = W_q· X_{trend} 
\end{equation}
\begin{equation}
    K = W_k ·X_{trend} 
\end{equation}
\begin{equation}
    V = W_v ·X_{trend}, 
\end{equation}
where, the projection matrices \(W_q, W_k, W_v \in \mathbb{R}^{d \times c}\) have dimension $d$ for the attention mechanism. Thereafter, EFA maps $Q$ and $K$ to the frequency domain using Fast Fourier Transform (FFT) for spectral decomposition, enabling element-wise complex multiplication to model their interaction. To mitigate numerical instability from this operation, L2 normalization is applied to the result.

For a one-dimensional time series $x(t)$, its Fourier transform is defined as:

\begin{equation}
\mathcal{F}\{x(t)\}(\omega) = \int_{-\infty}^{+\infty} x(t) e^{-i\omega t} \, dt,
\end{equation}
where $\omega$ denotes the angular frequency. This transformation decomposes the original signal into a superposition of sin/cos basis functions of different frequencies.

The interaction between $Q$ and $K$ is defined as:

\begin{equation}
\phi =   \frac{\mathcal{F}(Q) \odot \mathcal{F}(K)}{\|\mathcal{F}(Q) \odot \mathcal{F}(K)\|_2 + \epsilon},
\end{equation}
where $\mathcal{F}$ denotes the FFT, $\epsilon$ is an infinitesimal positive real number (typically on the order of $10^{-8}$), serving as a regularization term to prevent division by zero. \(\odot\) denotes element-wise multiplication. $\phi$ is the result of interaction between $Q$ and $K$. Assuming the frequency-domain representations of $Q$ and $K$ are the superposition of low-frequency component $f_{\text{low}}(\omega)$ and high-frequency component $f_{\text{high}}(\omega)$, the complex multiplication can be expressed as:

\begin{equation}
\begin{split}
\left[f_{\text{low}}(\omega) + f_{\text{high}}(\omega)\right] \odot \left[f_{\text{low}}(\omega) + f_{\text{high}}(\omega)\right] &= \\ f_{\text{low}}^2(\omega) + 2f_{\text{low}}(\omega)f_{\text{high}}(\omega) 
 + f_{\text{high}}^2(\omega),
\end{split}
\end{equation}

Since the value of low-frequency components typically dominates that of high-frequency components, the $f_{\text{low}}^2(\omega)$ term in the above equation becomes the primary contribution, enhancing the interaction effect of low-frequency trend.

After enhancing the low-frequency trend, EFA converting the frequency-domain signal back to the time domain via Inverse Fourier Transform (IFFT):

\begin{equation}
\widehat{\phi} = \mathcal{F}^{-1}(\phi),
\end{equation}
where $\mathcal{F}^{-1}$ is inverse FFT. At this point, the low-frequency trend components in the reconstructed signal are amplified due to frequency-domain weighting, while high-frequency fluctuation components are attenuated.

Then, the frequency-enhanced matrix $\mathbf{\widehat{\phi}} $ is multiplied element-wise with an all-ones matrix $\mathbb{1}$ via broadcasting:

\begin{equation}
    \widehat{\phi}' = \widehat{\phi} \otimes \mathbb{1},
\end{equation}
where $\otimes$ denotes the element-wise multiplication with broadcasting. The result $\widehat{\phi}'$ is then normalized by the square root of the number of dimensions and the attention weights are computed using an softmax activation function:

\begin{equation}
    \S = \text{softmax}(\frac{\widehat{\phi}'}{\sqrt{d}}),
\end{equation}
where $\S$ is the attention score and $d$ is the dimension of the attention head. 

Finally, the output is obtained by multiplying these weights with the exponentially activated value matrix $V$:

\begin{equation}
    \widehat{V} = \text{clamp}(\exp(V))
\end{equation}

\begin{equation}
    X_{trend}^{out} = \S \times \widehat{V},
\end{equation}
where $\text{clamp}$ denotes numerical truncation to prevent exponential explosion. $X_{trend}^{out}$ is the output of EFA.

\begin{figure}[h]
\centering
\includegraphics[width=1\linewidth]{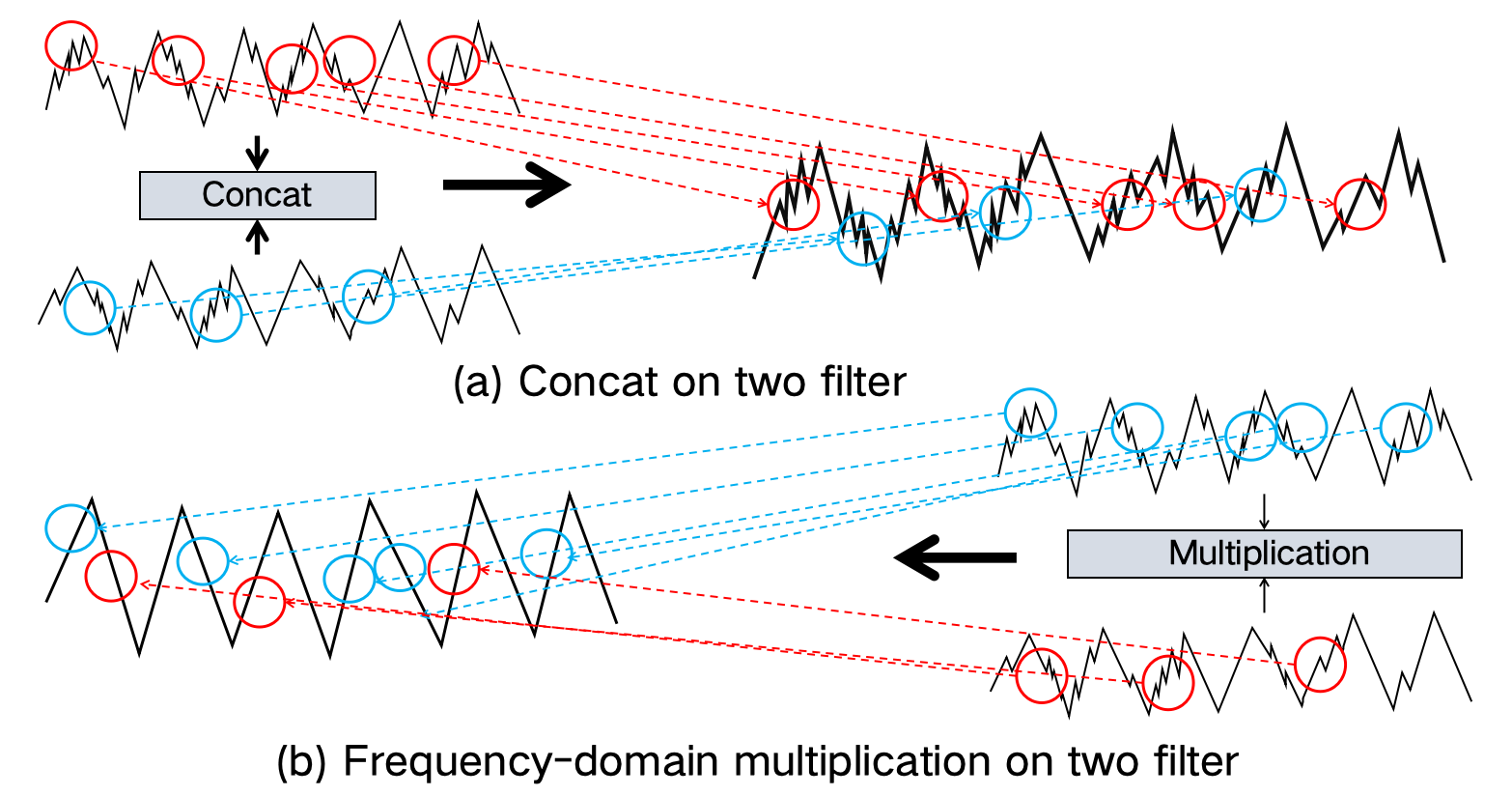}
\caption{Integration of PaiFilter and TexFilter. (a) Direct concatenation of two filters: Noise is superimposed when simply combining results; (b) Frequency-domain multiplication: Noise is filtered rather than aggregated.}
\label{fig:two filter}
\end{figure}

\subsection{CrossFilter Block}
Although EFA effectively captures low-frequency signals, it has limited capability in processing high-frequency signals in the seasonal component, which leads to poor robustness in modeling. To address this, we propose CrossFilter to enhance the modeling capability of high-frequency signals. This method skillfully combines two filters based on kernel functions of different scales. It incorporates modules designed to handle both complex and simple data, enabling excellent modeling performance across diverse datasets.

\textbf{PaiFilter (Plain Shaping Filter):}
The PaiFilter is a filtering mechanism designed to denoise the high-frequency signals.\cite{14} By including this method, the CrossFilter can help the XLinear model better on high-frequency signal. This filter employs a learnable generic frequency kernel with randomly initialized parameters to accomplish signal filtering. The filtering process is mathematically expressed as follows:
\begin{equation}
S_{pai}^{out} = \mathcal{F}^{-1}(\mathcal{F}(S) \odot \mathcal{S})
\end{equation}
where, the $S$ represents the input time series data, and the \(\mathcal{H}\) denotes the learnable frequency kernel. $S_{pai}^{out}$ is the series filtered by PaiFilter. The number of parameters of $\mathcal{H}$ is small, enabling faster modeling. However, due to the limited number of parameters, PaiFilter exhibits limitations when handling complex datasets.

\textbf{TexFilter (Contextual Shaping Filter):}
The TexFilter addresses the limitations of PaiFilter when applied to complex datasets. By employing TexFilter with PaiFilter, CrossFilter can work well on both simple and complex datasets. The TexFilter leverages a neural network $\mathcal{H}_{\varphi}$ to learn a frequency filter that aligns with the input signal characteristics.  Specifically, \(\mathcal{H}_{\varphi} = \sigma \prod_{i=1}^{n} \mathcal{W}_i\) is derived from a series of complex-value multiplications involving $n$ learnable parameters \(\mathcal{W}_{i}\). The network $\mathcal{H}_{\varphi}$ dynamically modulates filtering weights based on the input data, enabling adaptability to complex scenarios. The filtering process is mathematically expressed as follows:

\begin{equation}
S_{tex}^{out} = \mathcal{F}^{-1}(\mathcal{F}(S) \odot \mathcal{H}_{\varphi}(\mathcal{F}(S)))
\end{equation}
where, the \(\mathcal{H}_{\varphi}\) dynamically adjusts weights according to the input signal's spectrum, $S_{tex}^{out}$ is the output of TexFilter. Through these complex-value operations, \(\mathcal{H}_\varphi\) effectively accommodates intricate patterns and excels at capturing complex dependencies inherent in large-scale, complex datasets.

\textbf{Cross Integration:}
Given that directly fusing PaiFilter and TexFilter via concatenation would result in noise accumulation (as illustrated in Figure \ref{fig:two filter}.), we propose cross integration. In this method, we deploy TexFilter and PaiFilter in parallel branches to denoise the original season component $X_{season}$ independently. The filtered outputs $X_{tex}$ and $X_{pai}$ then undergo nonlinear transformation via the GELU (Gaussian Error Linear Unit) activation function, which employs an adaptive probability-weighted mechanism to suppress residual noise. Finally, the two filter outputs are fused via element-wise multiplication, formalized as:
\begin{equation}
    X_{pai} = \text{PaiFilter}(X_{season})
\end{equation}

\begin{equation}
    X_{tex} = \text{TexFilter}(X_{season})
\end{equation}

\begin{equation}
X_{season}^{out} = X_{tex} \odot \text{GELU}(X_{pai}) + X_{pai} \odot \text{GELU}(X_{tex}),
\end{equation}
where, the $X_{season}^{out}$ is the output of CrossFilter. This approach ensures mutual noise suppression during fusion, leveraging filter complementarity to enhance denoising efficiency.



\begin{table*}
\caption{Results with different prediction lengths $O \in \{96, 192, 336, 720\}$. The optimal results are highlighted in \color{red}{red}\color{black}, with the second-best performances denoted in \color{blue}blue\color{black}
. The lower MSE and MAE means better performance.}
\begin{tabular}{cccccccccccccccccc}
\hline
\multicolumn{2}{c}{ }                     & \multicolumn{2}{c}{XLinear}                                  & \multicolumn{2}{c}{TexFilter}                                                                                & \multicolumn{2}{c}{PaiFilter}                                                                               & \multicolumn{2}{c}{iTransformer}                                                                            & \multicolumn{2}{c}{PatchTST}                                                                                & \multicolumn{2}{c}{FEDformer}                                                                               & \multicolumn{2}{c}{TimesNet}                                                                                & \multicolumn{2}{c}{DLinear}                                                                                 \\
\multicolumn{2}{c}{Metrics}                    & MSE                          & MAE                          & MSE                                                   & MAE                                                  & MSE                                                  & MAE                                                  & MSE                                                  & MAE                                                  & MSE                                                  & MAE                                                  & MSE                                                  & MAE                                                  & MSE                                                  & MAE                                                  & MSE                                                  & MAE                                                  \\
\hline
                                 & 96          & {\color[HTML]{FE0000} 0.318} & {\color[HTML]{FE0000} 0.350} & \cellcolor[HTML]{FFFFFF}{\color[HTML]{000000} 0.321}  & \cellcolor[HTML]{FFFFFF}{\color[HTML]{000000} 0.361} & \cellcolor[HTML]{FFFFFF}{\color[HTML]{3531FF} 0.318} & \cellcolor[HTML]{FFFFFF}{\color[HTML]{3531FF} 0.358} & \cellcolor[HTML]{FFFFFF}{\color[HTML]{000000} 0.334} & \cellcolor[HTML]{FFFFFF}{\color[HTML]{000000} 0.368} & \cellcolor[HTML]{FFFFFF}{\color[HTML]{000000} 0.329} & \cellcolor[HTML]{FFFFFF}{\color[HTML]{000000} 0.367} & \cellcolor[HTML]{FFFFFF}{\color[HTML]{000000} 0.379} & \cellcolor[HTML]{FFFFFF}{\color[HTML]{000000} 0.419} & \cellcolor[HTML]{FFFFFF}{\color[HTML]{000000} 0.338} & \cellcolor[HTML]{FFFFFF}{\color[HTML]{000000} 0.375} & \cellcolor[HTML]{FFFFFF}{\color[HTML]{000000} 0.344} & \cellcolor[HTML]{FFFFFF}{\color[HTML]{000000} 0.370} \\
                                 & 192         & {\color[HTML]{FE0000} 0.361} & {\color[HTML]{FE0000} 0.375} & \cellcolor[HTML]{FFFFFF}{\color[HTML]{000000} 0.367}  & \cellcolor[HTML]{FFFFFF}{\color[HTML]{000000} 0.387} & \cellcolor[HTML]{FFFFFF}{\color[HTML]{3531FF} 0.364} & \cellcolor[HTML]{FFFFFF}{\color[HTML]{3531FF} 0.383} & \cellcolor[HTML]{FFFFFF}{\color[HTML]{000000} 0.377} & \cellcolor[HTML]{FFFFFF}{\color[HTML]{000000} 0.391} & \cellcolor[HTML]{FFFFFF}{\color[HTML]{000000} 0.367} & \cellcolor[HTML]{FFFFFF}{\color[HTML]{000000} 0.385} & \cellcolor[HTML]{FFFFFF}{\color[HTML]{000000} 0.426} & \cellcolor[HTML]{FFFFFF}{\color[HTML]{000000} 0.441} & \cellcolor[HTML]{FFFFFF}{\color[HTML]{000000} 0.374} & \cellcolor[HTML]{FFFFFF}{\color[HTML]{000000} 0.387} & \cellcolor[HTML]{FFFFFF}{\color[HTML]{000000} 0.379} & \cellcolor[HTML]{FFFFFF}{\color[HTML]{000000} 0.393} \\
                                 & 336         & {\color[HTML]{FE0000} 0.381} & {\color[HTML]{FE0000} 0.397} & \cellcolor[HTML]{FFFFFF}{\color[HTML]{000000} 0.401}  & \cellcolor[HTML]{FFFFFF}{\color[HTML]{000000} 0.409} & \cellcolor[HTML]{FFFFFF}{\color[HTML]{3531FF} 0.396} & \cellcolor[HTML]{FFFFFF}{\color[HTML]{3531FF} 0.406} & \cellcolor[HTML]{FFFFFF}{\color[HTML]{000000} 0.426} & \cellcolor[HTML]{FFFFFF}{\color[HTML]{000000} 0.420}  & \cellcolor[HTML]{FFFFFF}{\color[HTML]{000000} 0.399} & \cellcolor[HTML]{FFFFFF}{\color[HTML]{000000} 0.410}  & \cellcolor[HTML]{FFFFFF}{\color[HTML]{000000} 0.445} & \cellcolor[HTML]{FFFFFF}{\color[HTML]{000000} 0.459} & \cellcolor[HTML]{FFFFFF}{\color[HTML]{000000} 0.410}  & \cellcolor[HTML]{FFFFFF}{\color[HTML]{000000} 0.411} & \cellcolor[HTML]{FFFFFF}{\color[HTML]{000000} 0.410}  & \cellcolor[HTML]{FFFFFF}{\color[HTML]{000000} 0.411} \\
                                 & 720         & {\color[HTML]{FE0000} 0.443} & {\color[HTML]{FE0000} 0.432} & \cellcolor[HTML]{FFFFFF}{\color[HTML]{000000} 0.477}  & \cellcolor[HTML]{FFFFFF}{\color[HTML]{000000} 0.448} & \cellcolor[HTML]{FFFFFF}{\color[HTML]{3531FF} 0.456} & \cellcolor[HTML]{FFFFFF}{\color[HTML]{3531FF} 0.444} & \cellcolor[HTML]{FFFFFF}{\color[HTML]{000000} 0.491} & \cellcolor[HTML]{FFFFFF}{\color[HTML]{000000} 0.459} & \cellcolor[HTML]{FFFFFF}{\color[HTML]{000000} 0.454} & \cellcolor[HTML]{FFFFFF}{\color[HTML]{000000} 0.439} & \cellcolor[HTML]{FFFFFF}{\color[HTML]{000000} 0.543} & \cellcolor[HTML]{FFFFFF}{\color[HTML]{000000} 0.490}  & \cellcolor[HTML]{FFFFFF}{\color[HTML]{000000} 0.478} & \cellcolor[HTML]{FFFFFF}{\color[HTML]{000000} 0.450}  & \cellcolor[HTML]{FFFFFF}{\color[HTML]{000000} 0.473} & \cellcolor[HTML]{FFFFFF}{\color[HTML]{000000} 0.450}  \\
\multirow{-5}{*}{\rotatebox{90}{ETTm1}}          & Avg         & {\color[HTML]{FE0000} 0.373} & {\color[HTML]{FE0000} 0.388} & \cellcolor[HTML]{FFFFFF}{\color[HTML]{000000} 0.391}  & \cellcolor[HTML]{FFFFFF}{\color[HTML]{000000} 0.401} & \cellcolor[HTML]{FFFFFF}{\color[HTML]{3531FF} 0.383} & \cellcolor[HTML]{FFFFFF}{\color[HTML]{3531FF} 0.397} & \cellcolor[HTML]{FFFFFF}{\color[HTML]{000000} 0.407} & \cellcolor[HTML]{FFFFFF}{\color[HTML]{000000} 0.409} & \cellcolor[HTML]{FFFFFF}{\color[HTML]{000000} 0.387} & \cellcolor[HTML]{FFFFFF}{\color[HTML]{000000} 0.400} & \cellcolor[HTML]{FFFFFF}{\color[HTML]{000000} 0.448} & \cellcolor[HTML]{FFFFFF}{\color[HTML]{000000} 0.452} & \cellcolor[HTML]{FFFFFF}{\color[HTML]{000000} 0.400} & \cellcolor[HTML]{FFFFFF}{\color[HTML]{000000} 0.405} & \cellcolor[HTML]{FFFFFF}{\color[HTML]{000000} 0.401} & \cellcolor[HTML]{FFFFFF}{\color[HTML]{000000} 0.406} \\
\hline
                                 & 96          & {\color[HTML]{000000} 0.176} & {\color[HTML]{000000} 0.260} & \cellcolor[HTML]{FFFFFF}{\color[HTML]{3531FF} 0.175}  & \cellcolor[HTML]{FFFFFF}{\color[HTML]{3531FF} 0.258} & \cellcolor[HTML]{FFFFFF}{\color[HTML]{FE0000} 0.174} & \cellcolor[HTML]{FFFFFF}{\color[HTML]{FE0000} 0.257} & \cellcolor[HTML]{FFFFFF}{\color[HTML]{000000} 0.180} & \cellcolor[HTML]{FFFFFF}{\color[HTML]{000000} 0.264} & \cellcolor[HTML]{FFFFFF}{\color[HTML]{000000} 0.175} & \cellcolor[HTML]{FFFFFF}{\color[HTML]{000000} 0.259} & \cellcolor[HTML]{FFFFFF}{\color[HTML]{000000} 0.203} & \cellcolor[HTML]{FFFFFF}{\color[HTML]{000000} 0.287} & \cellcolor[HTML]{FFFFFF}{\color[HTML]{000000} 0.187} & \cellcolor[HTML]{FFFFFF}{\color[HTML]{000000} 0.267} & \cellcolor[HTML]{FFFFFF}{\color[HTML]{000000} 0.187} & \cellcolor[HTML]{FFFFFF}{\color[HTML]{000000} 0.281} \\
                                 & 192         & {\color[HTML]{FE0000} 0.237} & {\color[HTML]{FE0000} 0.302} & \cellcolor[HTML]{FFFFFF}{\color[HTML]{000000} 0.240}   & \cellcolor[HTML]{FFFFFF}{\color[HTML]{000000} 0.301} & \cellcolor[HTML]{FFFFFF}{\color[HTML]{3531FF} 0.240} & \cellcolor[HTML]{FFFFFF}{\color[HTML]{3531FF} 0.300} & \cellcolor[HTML]{FFFFFF}{\color[HTML]{000000} 0.250} & \cellcolor[HTML]{FFFFFF}{\color[HTML]{000000} 0.309} & \cellcolor[HTML]{FFFFFF}{\color[HTML]{000000} 0.241} & \cellcolor[HTML]{FFFFFF}{\color[HTML]{000000} 0.302} & \cellcolor[HTML]{FFFFFF}{\color[HTML]{000000} 0.269} & \cellcolor[HTML]{FFFFFF}{\color[HTML]{000000} 0.328} & \cellcolor[HTML]{FFFFFF}{\color[HTML]{000000} 0.249} & \cellcolor[HTML]{FFFFFF}{\color[HTML]{000000} 0.309} & \cellcolor[HTML]{FFFFFF}{\color[HTML]{000000} 0.272} & \cellcolor[HTML]{FFFFFF}{\color[HTML]{000000} 0.349} \\
                                 & 336         & {\color[HTML]{FE0000} 0.286} & {\color[HTML]{FE0000} 0.337} & \cellcolor[HTML]{FFFFFF}{\color[HTML]{000000} 0.311}  & \cellcolor[HTML]{FFFFFF}{\color[HTML]{000000} 0.347} & \cellcolor[HTML]{FFFFFF}{\color[HTML]{3531FF} 0.297} & \cellcolor[HTML]{FFFFFF}{\color[HTML]{3531FF} 0.339} & \cellcolor[HTML]{FFFFFF}{\color[HTML]{000000} 0.311} & \cellcolor[HTML]{FFFFFF}{\color[HTML]{000000} 0.348} & \cellcolor[HTML]{FFFFFF}{\color[HTML]{000000} 0.305} & \cellcolor[HTML]{FFFFFF}{\color[HTML]{000000} 0.343} & \cellcolor[HTML]{FFFFFF}{\color[HTML]{000000} 0.325} & \cellcolor[HTML]{FFFFFF}{\color[HTML]{000000} 0.366} & \cellcolor[HTML]{FFFFFF}{\color[HTML]{000000} 0.321} & \cellcolor[HTML]{FFFFFF}{\color[HTML]{000000} 0.351} & \cellcolor[HTML]{FFFFFF}{\color[HTML]{000000} 0.316} & \cellcolor[HTML]{FFFFFF}{\color[HTML]{000000} 0.372} \\
                                 & 720         & {\color[HTML]{FE0000} 0.368} & {\color[HTML]{FE0000} 0.389} & \cellcolor[HTML]{FFFFFF}{\color[HTML]{000000} 0.414}  & \cellcolor[HTML]{FFFFFF}{\color[HTML]{000000} 0.405} & \cellcolor[HTML]{FFFFFF}{\color[HTML]{3531FF} 0.392} & \cellcolor[HTML]{FFFFFF}{\color[HTML]{3531FF} 0.393} & \cellcolor[HTML]{FFFFFF}{\color[HTML]{000000} 0.412} & \cellcolor[HTML]{FFFFFF}{\color[HTML]{000000} 0.407} & \cellcolor[HTML]{FFFFFF}{\color[HTML]{000000} 0.402} & \cellcolor[HTML]{FFFFFF}{\color[HTML]{000000} 0.400} & \cellcolor[HTML]{FFFFFF}{\color[HTML]{000000} 0.421} & \cellcolor[HTML]{FFFFFF}{\color[HTML]{000000} 0.415} & \cellcolor[HTML]{FFFFFF}{\color[HTML]{000000} 0.408} & \cellcolor[HTML]{FFFFFF}{\color[HTML]{000000} 0.403} & \cellcolor[HTML]{FFFFFF}{\color[HTML]{000000} 0.452} & \cellcolor[HTML]{FFFFFF}{\color[HTML]{000000} 0.457} \\
\multirow{-5}{*}{\rotatebox{90}{ETTm2}}          & Avg         & {\color[HTML]{FE0000} 0.269} & {\color[HTML]{FE0000} 0.320} & \cellcolor[HTML]{FFFFFF}{\color[HTML]{000000} 0.285}  & \cellcolor[HTML]{FFFFFF}{\color[HTML]{000000} 0.327} & \cellcolor[HTML]{FFFFFF}{\color[HTML]{3531FF} 0.275} & \cellcolor[HTML]{FFFFFF}{\color[HTML]{3531FF} 0.322} & \cellcolor[HTML]{FFFFFF}{\color[HTML]{000000} 0.288} & \cellcolor[HTML]{FFFFFF}{\color[HTML]{000000} 0.332} & \cellcolor[HTML]{FFFFFF}{\color[HTML]{000000} 0.280} & \cellcolor[HTML]{FFFFFF}{\color[HTML]{000000} 0.326} & \cellcolor[HTML]{FFFFFF}{\color[HTML]{000000} 0.304} & \cellcolor[HTML]{FFFFFF}{\color[HTML]{000000} 0.349} & \cellcolor[HTML]{FFFFFF}{\color[HTML]{000000} 0.291} & \cellcolor[HTML]{FFFFFF}{\color[HTML]{000000} 0.332} & \cellcolor[HTML]{FFFFFF}{\color[HTML]{000000} 0.306} & \cellcolor[HTML]{FFFFFF}{\color[HTML]{000000} 0.364} \\
\hline
                                 & 96          & 0.393                        & 0.416                        & \cellcolor[HTML]{FFFFFF}{\color[HTML]{3531FF} 0.382}  & \cellcolor[HTML]{FFFFFF}{\color[HTML]{3531FF} 0.402} & \cellcolor[HTML]{FFFFFF}{\color[HTML]{FE0000} 0.375} & \cellcolor[HTML]{FFFFFF}{\color[HTML]{FE0000} 0.394} & \cellcolor[HTML]{FFFFFF}{\color[HTML]{000000} 0.386} & \cellcolor[HTML]{FFFFFF}{\color[HTML]{000000} 0.405} & \cellcolor[HTML]{FFFFFF}{\color[HTML]{000000} 0.414} & \cellcolor[HTML]{FFFFFF}{\color[HTML]{000000} 0.419} & \cellcolor[HTML]{FFFFFF}{\color[HTML]{000000} 0.376} & \cellcolor[HTML]{FFFFFF}{\color[HTML]{000000} 0.420} & \cellcolor[HTML]{FFFFFF}{\color[HTML]{000000} 0.384} & \cellcolor[HTML]{FFFFFF}{\color[HTML]{000000} 0.402} & \cellcolor[HTML]{FFFFFF}{\color[HTML]{000000} 0.383} & \cellcolor[HTML]{FFFFFF}{\color[HTML]{000000} 0.396} \\
                                 & 192         & {\color[HTML]{3531FF} 0.430} & {\color[HTML]{3531FF} 0.426} & \cellcolor[HTML]{FFFFFF}{\color[HTML]{FE0000} 0.430}   & \cellcolor[HTML]{FFFFFF}{\color[HTML]{FE0000} 0.429} & \cellcolor[HTML]{FFFFFF}{\color[HTML]{000000} 0.436} & \cellcolor[HTML]{FFFFFF}{\color[HTML]{000000} 0.422} & \cellcolor[HTML]{FFFFFF}{\color[HTML]{000000} 0.441} & \cellcolor[HTML]{FFFFFF}{\color[HTML]{000000} 0.436} & \cellcolor[HTML]{FFFFFF}{\color[HTML]{000000} 0.460} & \cellcolor[HTML]{FFFFFF}{\color[HTML]{000000} 0.445} & \cellcolor[HTML]{FFFFFF}{\color[HTML]{000000} 0.420} & \cellcolor[HTML]{FFFFFF}{\color[HTML]{000000} 0.448} & \cellcolor[HTML]{FFFFFF}{\color[HTML]{000000} 0.436} & \cellcolor[HTML]{FFFFFF}{\color[HTML]{000000} 0.429} & \cellcolor[HTML]{FFFFFF}{\color[HTML]{000000} 0.433} & \cellcolor[HTML]{FFFFFF}{\color[HTML]{000000} 0.426} \\
                                 & 336         & {\color[HTML]{FE0000} 0.467} & {\color[HTML]{FE0000} 0.443} & \cellcolor[HTML]{FFFFFF}{\color[HTML]{3166FF} 0.472}  & \cellcolor[HTML]{FFFFFF}{\color[HTML]{000000} 0.451} & \cellcolor[HTML]{FFFFFF}{\color[HTML]{000000} 0.476} & \cellcolor[HTML]{FFFFFF}{\color[HTML]{3531FF} 0.443} & \cellcolor[HTML]{FFFFFF}{\color[HTML]{000000} 0.487} & \cellcolor[HTML]{FFFFFF}{\color[HTML]{000000} 0.458} & \cellcolor[HTML]{FFFFFF}{\color[HTML]{000000} 0.501} & \cellcolor[HTML]{FFFFFF}{\color[HTML]{000000} 0.466} & \cellcolor[HTML]{FFFFFF}{\color[HTML]{000000} 0.459} & \cellcolor[HTML]{FFFFFF}{\color[HTML]{000000} 0.465} & \cellcolor[HTML]{FFFFFF}{\color[HTML]{000000} 0.491} & \cellcolor[HTML]{FFFFFF}{\color[HTML]{000000} 0.469} & \cellcolor[HTML]{FFFFFF}{\color[HTML]{000000} 0.479} & \cellcolor[HTML]{FFFFFF}{\color[HTML]{000000} 0.457} \\
                                 & 720         & {\color[HTML]{FE0000} 0.462} & {\color[HTML]{FE0000} 0.465} & \cellcolor[HTML]{FFFFFF}{\color[HTML]{000000} 0.481}  & \cellcolor[HTML]{FFFFFF}{\color[HTML]{000000} 0.473} & \cellcolor[HTML]{FFFFFF}{\color[HTML]{3531FF} 0.474} & \cellcolor[HTML]{FFFFFF}{\color[HTML]{3531FF} 0.469} & \cellcolor[HTML]{FFFFFF}{\color[HTML]{000000} 0.503} & \cellcolor[HTML]{FFFFFF}{\color[HTML]{000000} 0.491} & \cellcolor[HTML]{FFFFFF}{\color[HTML]{000000} 0.500} & \cellcolor[HTML]{FFFFFF}{\color[HTML]{000000} 0.488} & \cellcolor[HTML]{FFFFFF}{\color[HTML]{000000} 0.506} & \cellcolor[HTML]{FFFFFF}{\color[HTML]{000000} 0.507} & \cellcolor[HTML]{FFFFFF}{\color[HTML]{000000} 0.521} & \cellcolor[HTML]{FFFFFF}{\color[HTML]{000000} 0.500} & \cellcolor[HTML]{FFFFFF}{\color[HTML]{000000} 0.517} & \cellcolor[HTML]{FFFFFF}{\color[HTML]{000000} 0.513} \\
\multirow{-5}{*}{\rotatebox{90}{ETTh1}}          & Avg         & {\color[HTML]{FE0000} 0.438} & {\color[HTML]{FE0000} 0.437} & \cellcolor[HTML]{FFFFFF}{\color[HTML]{000000} 0.441}  & \cellcolor[HTML]{FFFFFF}{\color[HTML]{000000} 0.438} & \cellcolor[HTML]{FFFFFF}{\color[HTML]{3531FF} 0.440} & \cellcolor[HTML]{FFFFFF}{\color[HTML]{3531FF} 0.432} & \cellcolor[HTML]{FFFFFF}{\color[HTML]{000000} 0.454} & \cellcolor[HTML]{FFFFFF}{\color[HTML]{000000} 0.447} & \cellcolor[HTML]{FFFFFF}{\color[HTML]{000000} 0.468} & \cellcolor[HTML]{FFFFFF}{\color[HTML]{000000} 0.454} & \cellcolor[HTML]{FFFFFF}{\color[HTML]{000000} 0.440} & \cellcolor[HTML]{FFFFFF}{\color[HTML]{000000} 0.460} & \cellcolor[HTML]{FFFFFF}{\color[HTML]{000000} 0.458} & \cellcolor[HTML]{FFFFFF}{\color[HTML]{000000} 0.450} & \cellcolor[HTML]{FFFFFF}{\color[HTML]{000000} 0.453} & \cellcolor[HTML]{FFFFFF}{\color[HTML]{000000} 0.448} \\
\hline
                                 & 96 & 0.295                        & 0.351                        & \cellcolor[HTML]{FFFFFF}{\color[HTML]{3531FF} 0.293}  & \cellcolor[HTML]{FFFFFF}{\color[HTML]{3531FF} 0.343} & \cellcolor[HTML]{FFFFFF}{\color[HTML]{FE0000} 0.292} & \cellcolor[HTML]{FFFFFF}{\color[HTML]{FE0000} 0.343} & \cellcolor[HTML]{FFFFFF}{\color[HTML]{000000} 0.297} & \cellcolor[HTML]{FFFFFF}{\color[HTML]{000000} 0.349} & \cellcolor[HTML]{FFFFFF}{\color[HTML]{000000} 0.302} & \cellcolor[HTML]{FFFFFF}{\color[HTML]{000000} 0.348} & \cellcolor[HTML]{FFFFFF}{\color[HTML]{000000} 0.358} & \cellcolor[HTML]{FFFFFF}{\color[HTML]{000000} 0.397} & \cellcolor[HTML]{FFFFFF}{\color[HTML]{000000} 0.340} & \cellcolor[HTML]{FFFFFF}{\color[HTML]{000000} 0.374} & \cellcolor[HTML]{FFFFFF}{\color[HTML]{000000} 0.320} & \cellcolor[HTML]{FFFFFF}{\color[HTML]{000000} 0.374} \\
                                 & 192         & {\color[HTML]{FE0000} 0.365} & {\color[HTML]{FE0000} 0.389} & \cellcolor[HTML]{FFFFFF}{\color[HTML]{000000} 0.374}  & \cellcolor[HTML]{FFFFFF}{\color[HTML]{000000} 0.396} & \cellcolor[HTML]{FFFFFF}{\color[HTML]{3531FF} 0.369} & \cellcolor[HTML]{FFFFFF}{\color[HTML]{3531FF} 0.395} & \cellcolor[HTML]{FFFFFF}{\color[HTML]{000000} 0.380} & \cellcolor[HTML]{FFFFFF}{\color[HTML]{000000} 0.400} & \cellcolor[HTML]{FFFFFF}{\color[HTML]{000000} 0.388} & \cellcolor[HTML]{FFFFFF}{\color[HTML]{000000} 0.400} & \cellcolor[HTML]{FFFFFF}{\color[HTML]{000000} 0.429} & \cellcolor[HTML]{FFFFFF}{\color[HTML]{000000} 0.439} & \cellcolor[HTML]{FFFFFF}{\color[HTML]{000000} 0.402} & \cellcolor[HTML]{FFFFFF}{\color[HTML]{000000} 0.414} & \cellcolor[HTML]{FFFFFF}{\color[HTML]{000000} 0.449} & \cellcolor[HTML]{FFFFFF}{\color[HTML]{000000} 0.454} \\
                                 & 336         & {\color[HTML]{FE0000} 0.399} & {\color[HTML]{FE0000} 0.421} & \cellcolor[HTML]{FFFFFF}{\color[HTML]{3531FF} 0.417}  & \cellcolor[HTML]{FFFFFF}{\color[HTML]{3531FF} 0.430} & \cellcolor[HTML]{FFFFFF}{\color[HTML]{000000} 0.420} & \cellcolor[HTML]{FFFFFF}{\color[HTML]{000000} 0.432} & \cellcolor[HTML]{FFFFFF}{\color[HTML]{000000} 0.428} & \cellcolor[HTML]{FFFFFF}{\color[HTML]{000000} 0.432} & \cellcolor[HTML]{FFFFFF}{\color[HTML]{000000} 0.426} & \cellcolor[HTML]{FFFFFF}{\color[HTML]{000000} 0.433} & \cellcolor[HTML]{FFFFFF}{\color[HTML]{000000} 0.496} & \cellcolor[HTML]{FFFFFF}{\color[HTML]{000000} 0.487} & \cellcolor[HTML]{FFFFFF}{\color[HTML]{000000} 0.452} & \cellcolor[HTML]{FFFFFF}{\color[HTML]{000000} 0.452} & \cellcolor[HTML]{FFFFFF}{\color[HTML]{000000} 0.467} & \cellcolor[HTML]{FFFFFF}{\color[HTML]{000000} 0.469} \\
                                 & 720         & {\color[HTML]{FE0000} 0.419} & {\color[HTML]{FE0000} 0.440} & \cellcolor[HTML]{FFFFFF}{\color[HTML]{000000} 0.449}  & \cellcolor[HTML]{FFFFFF}{\color[HTML]{000000} 0.460} & \cellcolor[HTML]{FFFFFF}{\color[HTML]{3531FF} 0.430} & \cellcolor[HTML]{FFFFFF}{\color[HTML]{3531FF} 0.446} & \cellcolor[HTML]{FFFFFF}{\color[HTML]{000000} 0.427} & \cellcolor[HTML]{FFFFFF}{\color[HTML]{000000} 0.445} & \cellcolor[HTML]{FFFFFF}{\color[HTML]{000000} 0.431} & \cellcolor[HTML]{FFFFFF}{\color[HTML]{000000} 0.446} & \cellcolor[HTML]{FFFFFF}{\color[HTML]{000000} 0.463} & \cellcolor[HTML]{FFFFFF}{\color[HTML]{000000} 0.474} & \cellcolor[HTML]{FFFFFF}{\color[HTML]{000000} 0.462} & \cellcolor[HTML]{FFFFFF}{\color[HTML]{000000} 0.468} & \cellcolor[HTML]{FFFFFF}{\color[HTML]{000000} 0.656} & \cellcolor[HTML]{FFFFFF}{\color[HTML]{000000} 0.571} \\
\multirow{-5}{*}{\rotatebox{90}{ETTh2}} & Avg         & {\color[HTML]{FE0000} 0.369} & {\color[HTML]{FE0000} 0.400} & \cellcolor[HTML]{FFFFFF}{\color[HTML]{000000} 0.383}  & \cellcolor[HTML]{FFFFFF}{\color[HTML]{000000} 0.407} & \cellcolor[HTML]{FFFFFF}{\color[HTML]{3531FF} 0.377} & \cellcolor[HTML]{FFFFFF}{\color[HTML]{3531FF} 0.404} & \cellcolor[HTML]{FFFFFF}{\color[HTML]{000000} 0.383} & \cellcolor[HTML]{FFFFFF}{\color[HTML]{000000} 0.406} & \cellcolor[HTML]{FFFFFF}{\color[HTML]{000000} 0.386} & \cellcolor[HTML]{FFFFFF}{\color[HTML]{000000} 0.406} & \cellcolor[HTML]{FFFFFF}{\color[HTML]{000000} 0.436} & \cellcolor[HTML]{FFFFFF}{\color[HTML]{000000} 0.449} & \cellcolor[HTML]{FFFFFF}{\color[HTML]{000000} 0.414} & \cellcolor[HTML]{FFFFFF}{\color[HTML]{000000} 0.427} & \cellcolor[HTML]{FFFFFF}{\color[HTML]{000000} 0.473} & \cellcolor[HTML]{FFFFFF}{\color[HTML]{000000} 0.467} \\
\hline
                                 & 96          & {\color[HTML]{3531FF} 0.428} & {\color[HTML]{3531FF} 0.286} & \cellcolor[HTML]{FFFFFF}{\color[HTML]{000000} 0.430}  & \cellcolor[HTML]{FFFFFF}{\color[HTML]{000000} 0.294} & \cellcolor[HTML]{FFFFFF}{\color[HTML]{000000} 0.506} & \cellcolor[HTML]{FFFFFF}{\color[HTML]{000000} 0.336} & \cellcolor[HTML]{FFFFFF}{\color[HTML]{FE0000} 0.395} & \cellcolor[HTML]{FFFFFF}{\color[HTML]{FE0000} 0.268} & \cellcolor[HTML]{FFFFFF}{\color[HTML]{000000} 0.462} & \cellcolor[HTML]{FFFFFF}{\color[HTML]{000000} 0.295} & \cellcolor[HTML]{FFFFFF}{\color[HTML]{000000} 0.587} & \cellcolor[HTML]{FFFFFF}{\color[HTML]{000000} 0.366} & \cellcolor[HTML]{FFFFFF}{\color[HTML]{000000} 0.593} & \cellcolor[HTML]{FFFFFF}{\color[HTML]{000000} 0.321} & \cellcolor[HTML]{FFFFFF}{\color[HTML]{000000} 0.65}  & \cellcolor[HTML]{FFFFFF}{\color[HTML]{000000} 0.397} \\
                                 & 192         & {\color[HTML]{3531FF} 0.448} & {\color[HTML]{3531FF} 0.295} & \cellcolor[HTML]{FFFFFF}{\color[HTML]{000000} 0.452}  & \cellcolor[HTML]{FFFFFF}{\color[HTML]{000000} 0.307} & \cellcolor[HTML]{FFFFFF}{\color[HTML]{000000} 0.508} & \cellcolor[HTML]{FFFFFF}{\color[HTML]{000000} 0.333} & \cellcolor[HTML]{FFFFFF}{\color[HTML]{FE0000} 0.417} & \cellcolor[HTML]{FFFFFF}{\color[HTML]{FE0000} 0.276} & \cellcolor[HTML]{FFFFFF}{\color[HTML]{000000} 0.466} & \cellcolor[HTML]{FFFFFF}{\color[HTML]{000000} 0.296} & \cellcolor[HTML]{FFFFFF}{\color[HTML]{000000} 0.604} & \cellcolor[HTML]{FFFFFF}{\color[HTML]{000000} 0.373} & \cellcolor[HTML]{FFFFFF}{\color[HTML]{000000} 0.617} & \cellcolor[HTML]{FFFFFF}{\color[HTML]{000000} 0.336} & \cellcolor[HTML]{FFFFFF}{\color[HTML]{000000} 0.600} & \cellcolor[HTML]{FFFFFF}{\color[HTML]{000000} 0.372} \\
                                 & 336         & {\color[HTML]{3531FF} 0.460} & {\color[HTML]{3531FF} 0.299} & \cellcolor[HTML]{FFFFFF}{\color[HTML]{000000} 0.470}  & \cellcolor[HTML]{FFFFFF}{\color[HTML]{000000} 0.316} & \cellcolor[HTML]{FFFFFF}{\color[HTML]{000000} 0.518} & \cellcolor[HTML]{FFFFFF}{\color[HTML]{000000} 0.335} & \cellcolor[HTML]{FFFFFF}{\color[HTML]{FE0000} 0.433} & \cellcolor[HTML]{FFFFFF}{\color[HTML]{FE0000} 0.283} & \cellcolor[HTML]{FFFFFF}{\color[HTML]{000000} 0.482} & \cellcolor[HTML]{FFFFFF}{\color[HTML]{000000} 0.304} & \cellcolor[HTML]{FFFFFF}{\color[HTML]{000000} 0.621} & \cellcolor[HTML]{FFFFFF}{\color[HTML]{000000} 0.383} & \cellcolor[HTML]{FFFFFF}{\color[HTML]{000000} 0.629} & \cellcolor[HTML]{FFFFFF}{\color[HTML]{000000} 0.336} & \cellcolor[HTML]{FFFFFF}{\color[HTML]{000000} 0.606} & \cellcolor[HTML]{FFFFFF}{\color[HTML]{000000} 0.374} \\
                                 & 720         & {\color[HTML]{3531FF} 0.478} & {\color[HTML]{3531FF} 0.316} & \cellcolor[HTML]{FFFFFF}{\color[HTML]{000000} 0.498}  & \cellcolor[HTML]{FFFFFF}{\color[HTML]{000000} 0.323} & \cellcolor[HTML]{FFFFFF}{\color[HTML]{000000} 0.553} & \cellcolor[HTML]{FFFFFF}{\color[HTML]{000000} 0.354} & \cellcolor[HTML]{FFFFFF}{\color[HTML]{FE0000} 0.467} & \cellcolor[HTML]{FFFFFF}{\color[HTML]{FE0000} 0.302} & \cellcolor[HTML]{FFFFFF}{\color[HTML]{000000} 0.514} & \cellcolor[HTML]{FFFFFF}{\color[HTML]{000000} 0.322} & \cellcolor[HTML]{FFFFFF}{\color[HTML]{000000} 0.626} & \cellcolor[HTML]{FFFFFF}{\color[HTML]{000000} 0.382} & \cellcolor[HTML]{FFFFFF}{\color[HTML]{000000} 0.640} & \cellcolor[HTML]{FFFFFF}{\color[HTML]{000000} 0.350} & \cellcolor[HTML]{FFFFFF}{\color[HTML]{000000} 0.646} & \cellcolor[HTML]{FFFFFF}{\color[HTML]{000000} 0.395} \\
\multirow{-5}{*}{\rotatebox{90}{Traffic}}        & Avg         & {\color[HTML]{3611FF} 0.453} & {\color[HTML]{3531FF} 0.299} & \cellcolor[HTML]{FFFFFF}{\color[HTML]{000000} 0.462} & \cellcolor[HTML]{FFFFFF}{\color[HTML]{000000} 0.310} & \cellcolor[HTML]{FFFFFF}{\color[HTML]{000000} 0.521} & \cellcolor[HTML]{FFFFFF}{\color[HTML]{000000} 0.339} & \cellcolor[HTML]{FFFFFF}{\color[HTML]{FE0000} 0.428} & \cellcolor[HTML]{FFFFFF}{\color[HTML]{FE0000} 0.282} & \cellcolor[HTML]{FFFFFF}{\color[HTML]{000000} 0.481} & \cellcolor[HTML]{FFFFFF}{\color[HTML]{000000} 0.304} & \cellcolor[HTML]{FFFFFF}{\color[HTML]{000000} 0.609} & \cellcolor[HTML]{FFFFFF}{\color[HTML]{000000} 0.376} & \cellcolor[HTML]{FFFFFF}{\color[HTML]{000000} 0.619} & \cellcolor[HTML]{FFFFFF}{\color[HTML]{000000} 0.335} & \cellcolor[HTML]{FFFFFF}{\color[HTML]{000000} 0.625} & \cellcolor[HTML]{FFFFFF}{\color[HTML]{000000} 0.384} \\
\hline
                                 & 96          & {\color[HTML]{FE0000} 0.081} & {\color[HTML]{FE0000} 0.197} & \cellcolor[HTML]{FFFFFF}{\color[HTML]{000000} 0.091}  & \cellcolor[HTML]{FFFFFF}{\color[HTML]{000000} 0.211} & \cellcolor[HTML]{FFFFFF}{\color[HTML]{3531FF} 0.083} & \cellcolor[HTML]{FFFFFF}{\color[HTML]{3531FF} 0.202} & \cellcolor[HTML]{FFFFFF}{\color[HTML]{000000} 0.086} & \cellcolor[HTML]{FFFFFF}{\color[HTML]{000000} 0.206} & \cellcolor[HTML]{FFFFFF}{\color[HTML]{330001} 0.088} & \cellcolor[HTML]{FFFFFF}{\color[HTML]{330001} 0.205} & \cellcolor[HTML]{FFFFFF}{\color[HTML]{000000} 0.148} & \cellcolor[HTML]{FFFFFF}{\color[HTML]{000000} 0.278} & \cellcolor[HTML]{FFFFFF}{\color[HTML]{000000} 0.107} & \cellcolor[HTML]{FFFFFF}{\color[HTML]{000000} 0.234} & \cellcolor[HTML]{FFFFFF}{\color[HTML]{000000} 0.085} & \cellcolor[HTML]{FFFFFF}{\color[HTML]{000000} 0.210} \\
                                 & 192         & {\color[HTML]{FE0000} 0.170} & {\color[HTML]{FE0000} 0.291} & \cellcolor[HTML]{FFFFFF}{\color[HTML]{000000} 0.186}  & \cellcolor[HTML]{FFFFFF}{\color[HTML]{000000} 0.305} & \cellcolor[HTML]{FFFFFF}{\color[HTML]{3531FF} 0.174} & \cellcolor[HTML]{FFFFFF}{\color[HTML]{3531FF} 0.296} & \cellcolor[HTML]{FFFFFF}{\color[HTML]{000000} 0.177} & \cellcolor[HTML]{FFFFFF}{\color[HTML]{000000} 0.299} & \cellcolor[HTML]{FFFFFF}{\color[HTML]{000000} 0.176} & \cellcolor[HTML]{FFFFFF}{\color[HTML]{000000} 0.299} & \cellcolor[HTML]{FFFFFF}{\color[HTML]{000000} 0.271} & \cellcolor[HTML]{FFFFFF}{\color[HTML]{000000} 0.315} & \cellcolor[HTML]{FFFFFF}{\color[HTML]{000000} 0.226} & \cellcolor[HTML]{FFFFFF}{\color[HTML]{000000} 0.344} & \cellcolor[HTML]{FFFFFF}{\color[HTML]{000000} 0.178} & \cellcolor[HTML]{FFFFFF}{\color[HTML]{000000} 0.299} \\
                                 & 336         & {\color[HTML]{3531FF} 0.319} & {\color[HTML]{3531FF} 0.407} & \cellcolor[HTML]{FFFFFF}{\color[HTML]{000000} 0.380}   & \cellcolor[HTML]{FFFFFF}{\color[HTML]{000000} 0.449} & \cellcolor[HTML]{FFFFFF}{\color[HTML]{000000} 0.326} & \cellcolor[HTML]{FFFFFF}{\color[HTML]{000000} 0.413} & \cellcolor[HTML]{FFFFFF}{\color[HTML]{000000} 0.331} & \cellcolor[HTML]{FFFFFF}{\color[HTML]{000000} 0.417} & \cellcolor[HTML]{FFFFFF}{\color[HTML]{FE0000} 0.301} & \cellcolor[HTML]{FFFFFF}{\color[HTML]{FE0000} 0.397} & \cellcolor[HTML]{FFFFFF}{\color[HTML]{000000} 0.460} & \cellcolor[HTML]{FFFFFF}{\color[HTML]{000000} 0.427} & \cellcolor[HTML]{FFFFFF}{\color[HTML]{000000} 0.367} & \cellcolor[HTML]{FFFFFF}{\color[HTML]{000000} 0.448} & \cellcolor[HTML]{FFFFFF}{\color[HTML]{000000} 0.298} & \cellcolor[HTML]{FFFFFF}{\color[HTML]{000000} 0.409} \\
                                 & 720         & {\color[HTML]{FE0000} 0.813} & {\color[HTML]{FE0000} 0.677} & \cellcolor[HTML]{FFFFFF}{\color[HTML]{000000} 0.896}  & \cellcolor[HTML]{FFFFFF}{\color[HTML]{000000} 0.712} & \cellcolor[HTML]{FFFFFF}{\color[HTML]{3531FF} 0.840} & \cellcolor[HTML]{FFFFFF}{\color[HTML]{3531FF} 0.670} & \cellcolor[HTML]{FFFFFF}{\color[HTML]{000000} 0.847} & \cellcolor[HTML]{FFFFFF}{\color[HTML]{000000} 0.691} & \cellcolor[HTML]{FFFFFF}{\color[HTML]{000000} 0.901} & \cellcolor[HTML]{FFFFFF}{\color[HTML]{000000} 0.714} & \cellcolor[HTML]{FFFFFF}{\color[HTML]{000000} 1.195} & \cellcolor[HTML]{FFFFFF}{\color[HTML]{000000} 0.695} & \cellcolor[HTML]{FFFFFF}{\color[HTML]{000000} 0.964} & \cellcolor[HTML]{FFFFFF}{\color[HTML]{000000} 0.746} & \cellcolor[HTML]{FFFFFF}{\color[HTML]{000000} 0.861} & \cellcolor[HTML]{FFFFFF}{\color[HTML]{000000} 0.671} \\
\multirow{-5}{*}{\rotatebox{90}{Exchange}}       & Avg         & {\color[HTML]{FE0000} 0.345} & {\color[HTML]{FE0000} 0.393} & \cellcolor[HTML]{FFFFFF}{\color[HTML]{000000} 0.388}  & \cellcolor[HTML]{FFFFFF}{\color[HTML]{000000} 0.419} & \cellcolor[HTML]{FFFFFF}{\color[HTML]{3531FF} 0.355} & \cellcolor[HTML]{FFFFFF}{\color[HTML]{3531FF} 0.395} & \cellcolor[HTML]{FFFFFF}{\color[HTML]{000000} 0.360} & \cellcolor[HTML]{FFFFFF}{\color[HTML]{000000} 0.403} & \cellcolor[HTML]{FFFFFF}{\color[HTML]{000000} 0.366} & \cellcolor[HTML]{FFFFFF}{\color[HTML]{000000} 0.403} & \cellcolor[HTML]{FFFFFF}{\color[HTML]{000000} 0.518} & \cellcolor[HTML]{FFFFFF}{\color[HTML]{000000} 0.428} & \cellcolor[HTML]{FFFFFF}{\color[HTML]{000000} 0.416} & \cellcolor[HTML]{FFFFFF}{\color[HTML]{000000} 0.443} & \cellcolor[HTML]{FFFFFF}{\color[HTML]{000000} 0.355} & \cellcolor[HTML]{FFFFFF}{\color[HTML]{000000} 0.397} \\
\hline
                                 & 96          & {\color[HTML]{FE0000} 0.159} & {\color[HTML]{FE0000} 0.204} & \cellcolor[HTML]{FFFFFF}{\color[HTML]{3531FF} 0.162}  & \cellcolor[HTML]{FFFFFF}{\color[HTML]{3531FF} 0.207} & \cellcolor[HTML]{FFFFFF}{\color[HTML]{000000} 0.164} & \cellcolor[HTML]{FFFFFF}{\color[HTML]{000000} 0.210} & \cellcolor[HTML]{FFFFFF}{\color[HTML]{000000} 0.174} & \cellcolor[HTML]{FFFFFF}{\color[HTML]{000000} 0.214} & \cellcolor[HTML]{FFFFFF}{\color[HTML]{000000} 0.177} & \cellcolor[HTML]{FFFFFF}{\color[HTML]{000000} 0.218} & \cellcolor[HTML]{FFFFFF}{\color[HTML]{000000} 0.217} & \cellcolor[HTML]{FFFFFF}{\color[HTML]{000000} 0.296} & \cellcolor[HTML]{FFFFFF}{\color[HTML]{000000} 0.172} & \cellcolor[HTML]{FFFFFF}{\color[HTML]{000000} 0.220} & \cellcolor[HTML]{FFFFFF}{\color[HTML]{000000} 0.194} & \cellcolor[HTML]{FFFFFF}{\color[HTML]{000000} 0.248} \\
                                 & 192         & {\color[HTML]{FE0000} 0.205} & {\color[HTML]{FE0000} 0.243} & \cellcolor[HTML]{FFFFFF}{\color[HTML]{3531FF} 0.210}  & \cellcolor[HTML]{FFFFFF}{\color[HTML]{3531FF} 0.250} & \cellcolor[HTML]{FFFFFF}{\color[HTML]{000000} 0.214} & \cellcolor[HTML]{FFFFFF}{\color[HTML]{000000} 0.252} & \cellcolor[HTML]{FFFFFF}{\color[HTML]{000000} 0.221} & \cellcolor[HTML]{FFFFFF}{\color[HTML]{000000} 0.254} & \cellcolor[HTML]{FFFFFF}{\color[HTML]{000000} 0.225} & \cellcolor[HTML]{FFFFFF}{\color[HTML]{000000} 0.259} & \cellcolor[HTML]{FFFFFF}{\color[HTML]{000000} 0.276} & \cellcolor[HTML]{FFFFFF}{\color[HTML]{000000} 0.336} & \cellcolor[HTML]{FFFFFF}{\color[HTML]{000000} 0.219} & \cellcolor[HTML]{FFFFFF}{\color[HTML]{000000} 0.261} & \cellcolor[HTML]{FFFFFF}{\color[HTML]{000000} 0.234} & \cellcolor[HTML]{FFFFFF}{\color[HTML]{000000} 0.290} \\
                                 & 336         & {\color[HTML]{FE0000} 0.256} & {\color[HTML]{FE0000} 0.283} & \cellcolor[HTML]{FFFFFF}{\color[HTML]{3531FF} 0.265}  & \cellcolor[HTML]{FFFFFF}{\color[HTML]{3531FF} 0.290} & \cellcolor[HTML]{FFFFFF}{\color[HTML]{000000} 0.268} & \cellcolor[HTML]{FFFFFF}{\color[HTML]{000000} 0.293} & \cellcolor[HTML]{FFFFFF}{\color[HTML]{000000} 0.278} & \cellcolor[HTML]{FFFFFF}{\color[HTML]{000000} 0.296} & \cellcolor[HTML]{FFFFFF}{\color[HTML]{000000} 0.278} & \cellcolor[HTML]{FFFFFF}{\color[HTML]{000000} 0.297} & \cellcolor[HTML]{FFFFFF}{\color[HTML]{000000} 0.339} & \cellcolor[HTML]{FFFFFF}{\color[HTML]{000000} 0.380} & \cellcolor[HTML]{FFFFFF}{\color[HTML]{000000} 0.280} & \cellcolor[HTML]{FFFFFF}{\color[HTML]{000000} 0.306} & \cellcolor[HTML]{FFFFFF}{\color[HTML]{000000} 0.283} & \cellcolor[HTML]{FFFFFF}{\color[HTML]{000000} 0.335} \\
                                 & 720         & {\color[HTML]{FE0000} 0.323} & {\color[HTML]{FE0000} 0.335} & \cellcolor[HTML]{FFFFFF}{\color[HTML]{3531FF} 0.342}  & \cellcolor[HTML]{FFFFFF}{\color[HTML]{3531FF} 0.340} & \cellcolor[HTML]{FFFFFF}{\color[HTML]{000000} 0.344} & \cellcolor[HTML]{FFFFFF}{\color[HTML]{000000} 0.342} & \cellcolor[HTML]{FFFFFF}{\color[HTML]{000000} 0.358} & \cellcolor[HTML]{FFFFFF}{\color[HTML]{000000} 0.347} & \cellcolor[HTML]{FFFFFF}{\color[HTML]{000000} 0.354} & \cellcolor[HTML]{FFFFFF}{\color[HTML]{000000} 0.348} & \cellcolor[HTML]{FFFFFF}{\color[HTML]{000000} 0.403} & \cellcolor[HTML]{FFFFFF}{\color[HTML]{000000} 0.428} & \cellcolor[HTML]{FFFFFF}{\color[HTML]{000000} 0.365} & \cellcolor[HTML]{FFFFFF}{\color[HTML]{000000} 0.359} & \cellcolor[HTML]{FFFFFF}{\color[HTML]{000000} 0.348} & \cellcolor[HTML]{FFFFFF}{\color[HTML]{000000} 0.385} \\
\multirow{-5}{*}{\rotatebox{90}{Weather}}        & Avg         & {\color[HTML]{FE0000} 0.238} & {\color[HTML]{FE0000} 0.265} & \cellcolor[HTML]{FFFFFF}{\color[HTML]{3531FF} 0.244}  & \cellcolor[HTML]{FFFFFF}{\color[HTML]{3531FF} 0.271} & \cellcolor[HTML]{FFFFFF}{\color[HTML]{000000} 0.247} & \cellcolor[HTML]{FFFFFF}{\color[HTML]{000000} 0.274} & \cellcolor[HTML]{FFFFFF}{\color[HTML]{000000} 0.257} & \cellcolor[HTML]{FFFFFF}{\color[HTML]{000000} 0.277} & \cellcolor[HTML]{FFFFFF}{\color[HTML]{000000} 0.258} & \cellcolor[HTML]{FFFFFF}{\color[HTML]{000000} 0.280} & \cellcolor[HTML]{FFFFFF}{\color[HTML]{000000} 0.308} & \cellcolor[HTML]{FFFFFF}{\color[HTML]{000000} 0.360} & \cellcolor[HTML]{FFFFFF}{\color[HTML]{000000} 0.259} & \cellcolor[HTML]{FFFFFF}{\color[HTML]{000000} 0.286} & \cellcolor[HTML]{FFFFFF}{\color[HTML]{000000} 0.264} & \cellcolor[HTML]{FFFFFF}{\color[HTML]{000000} 0.314} \\
\hline
                                 & 96          & {\color[HTML]{FE0000} 0.147} & {\color[HTML]{FE0000} 0.246} & \cellcolor[HTML]{FFFFFF}{\color[HTML]{3531FF} 0.147}  & \cellcolor[HTML]{FFFFFF}{\color[HTML]{3531FF} 0.245} & \cellcolor[HTML]{FFFFFF}{\color[HTML]{000000} 0.176} & \cellcolor[HTML]{FFFFFF}{\color[HTML]{000000} 0.264} & \cellcolor[HTML]{FFFFFF}{\color[HTML]{000000} 0.148} & \cellcolor[HTML]{FFFFFF}{\color[HTML]{000000} 0.240} & \cellcolor[HTML]{FFFFFF}{\color[HTML]{000000} 0.181} & \cellcolor[HTML]{FFFFFF}{\color[HTML]{000000} 0.270} & \cellcolor[HTML]{FFFFFF}{\color[HTML]{000000} 0.193} & \cellcolor[HTML]{FFFFFF}{\color[HTML]{000000} 0.308} & \cellcolor[HTML]{FFFFFF}{\color[HTML]{000000} 0.168} & \cellcolor[HTML]{FFFFFF}{\color[HTML]{000000} 0.272} & \cellcolor[HTML]{FFFFFF}{\color[HTML]{000000} 0.195} & \cellcolor[HTML]{FFFFFF}{\color[HTML]{000000} 0.277} \\
                                 & 192         & {\color[HTML]{FE0000} 0.155} & {\color[HTML]{FE0000} 0.249} & \cellcolor[HTML]{FFFFFF}{\color[HTML]{3531FF} 0.160}  & \cellcolor[HTML]{FFFFFF}{\color[HTML]{3531FF} 0.250} & \cellcolor[HTML]{FFFFFF}{\color[HTML]{000000} 0.185} & \cellcolor[HTML]{FFFFFF}{\color[HTML]{000000} 0.270} & \cellcolor[HTML]{FFFFFF}{\color[HTML]{000000} 0.162} & \cellcolor[HTML]{FFFFFF}{\color[HTML]{000000} 0.253} & \cellcolor[HTML]{FFFFFF}{\color[HTML]{000000} 0.188} & \cellcolor[HTML]{FFFFFF}{\color[HTML]{000000} 0.274} & \cellcolor[HTML]{FFFFFF}{\color[HTML]{000000} 0.201} & \cellcolor[HTML]{FFFFFF}{\color[HTML]{000000} 0.315} & \cellcolor[HTML]{FFFFFF}{\color[HTML]{000000} 0.184} & \cellcolor[HTML]{FFFFFF}{\color[HTML]{000000} 0.289} & \cellcolor[HTML]{FFFFFF}{\color[HTML]{000000} 0.194} & \cellcolor[HTML]{FFFFFF}{\color[HTML]{000000} 0.280} \\
                                 & 336         & {\color[HTML]{FE0000} 0.168} & {\color[HTML]{FE0000} 0.265} & \cellcolor[HTML]{FFFFFF}{\color[HTML]{3531FF} 0.173}  & \cellcolor[HTML]{FFFFFF}{\color[HTML]{3531FF} 0.267} & \cellcolor[HTML]{FFFFFF}{\color[HTML]{000000} 0.202} & \cellcolor[HTML]{FFFFFF}{\color[HTML]{000000} 0.286} & \cellcolor[HTML]{FFFFFF}{\color[HTML]{000000} 0.178} & \cellcolor[HTML]{FFFFFF}{\color[HTML]{000000} 0.269} & \cellcolor[HTML]{FFFFFF}{\color[HTML]{000000} 0.204} & \cellcolor[HTML]{FFFFFF}{\color[HTML]{000000} 0.293} & \cellcolor[HTML]{FFFFFF}{\color[HTML]{000000} 0.214} & \cellcolor[HTML]{FFFFFF}{\color[HTML]{000000} 0.329} & \cellcolor[HTML]{FFFFFF}{\color[HTML]{000000} 0.198} & \cellcolor[HTML]{FFFFFF}{\color[HTML]{000000} 0.300} & \cellcolor[HTML]{FFFFFF}{\color[HTML]{000000} 0.207} & \cellcolor[HTML]{FFFFFF}{\color[HTML]{000000} 0.296} \\
                                 & 720         & {\color[HTML]{FE0000} 0.196} & {\color[HTML]{FE0000} 0.290} & \cellcolor[HTML]{FFFFFF}{\color[HTML]{3531FF} 0.210}  & \cellcolor[HTML]{FFFFFF}{\color[HTML]{3531FF} 0.309} & \cellcolor[HTML]{FFFFFF}{\color[HTML]{000000} 0.242} & \cellcolor[HTML]{FFFFFF}{\color[HTML]{000000} 0.319} & \cellcolor[HTML]{FFFFFF}{\color[HTML]{000000} 0.225} & \cellcolor[HTML]{FFFFFF}{\color[HTML]{000000} 0.317} & \cellcolor[HTML]{FFFFFF}{\color[HTML]{000000} 0.246} & \cellcolor[HTML]{FFFFFF}{\color[HTML]{000000} 0.324} & \cellcolor[HTML]{FFFFFF}{\color[HTML]{000000} 0.246} & \cellcolor[HTML]{FFFFFF}{\color[HTML]{000000} 0.355} & \cellcolor[HTML]{FFFFFF}{\color[HTML]{000000} 0.220} & \cellcolor[HTML]{FFFFFF}{\color[HTML]{000000} 0.320} & \cellcolor[HTML]{FFFFFF}{\color[HTML]{000000} 0.242} & \cellcolor[HTML]{FFFFFF}{\color[HTML]{000000} 0.329} \\
\multirow{-5}{*}{\rotatebox{90}{ECL}}            & Avg         & {\color[HTML]{FE0000} 0.166} & {\color[HTML]{FE0000} 0.260} & \cellcolor[HTML]{FFFFFF}{\color[HTML]{3531FF} 0.172}  & \cellcolor[HTML]{FFFFFF}{\color[HTML]{3531FF} 0.267} & \cellcolor[HTML]{FFFFFF}{\color[HTML]{000000} 0.201} & \cellcolor[HTML]{FFFFFF}{\color[HTML]{000000} 0.284} & \cellcolor[HTML]{FFFFFF}{\color[HTML]{000000} 0.178} & \cellcolor[HTML]{FFFFFF}{\color[HTML]{000000} 0.269} & \cellcolor[HTML]{FFFFFF}{\color[HTML]{000000} 0.204} & \cellcolor[HTML]{FFFFFF}{\color[HTML]{000000} 0.290} & \cellcolor[HTML]{FFFFFF}{\color[HTML]{000000} 0.213} & \cellcolor[HTML]{FFFFFF}{\color[HTML]{000000} 0.326} & \cellcolor[HTML]{FFFFFF}{\color[HTML]{000000} 0.192} & \cellcolor[HTML]{FFFFFF}{\color[HTML]{000000} 0.295} & \cellcolor[HTML]{FFFFFF}{\color[HTML]{000000} 0.209} & \cellcolor[HTML]{FFFFFF}{\color[HTML]{000000} 0.295}\\
\hline
\end{tabular}
\label{tab:main}
\end{table*}

\section{EXPERIMENTS}
\subsection{Experimental Setup}
\textbf{Datasets:}  We conducted experiments on 8 public datasets tailored for assessing time series forecasting accuracy, spanning domains of energy, finance, and transportation. These datasets, widely adopted as benchmarking standards, exhibit diverse characteristics to systematically evaluate model generalization: scale ranges from large-scale collections (e.g., Weather, ETTm1) to small-scale counterparts (e.g., Exchange), while input dimensions vary from low-dimensional contexts (e.g., ETT, Weather) to high-dimensional ones (e.g., ECL, Traffic). Such diversity facilitates rigorous assessment of a model’s generalization ability across heterogeneous data distributions.

\textbf{Implement Details:} The experiments were performed on a single A800 80G GPU, utilizing an L2 loss function in conjunction with the Adam optimizer. We integrated RevIN as the instance normalization block.\cite{18} Additionally, hyperparameters including batch size and learning rate were meticulously optimized on the validation set, with optimal configurations selected based on performance. The batch size was tuned across $\{32, 64, 128, 256\}$ and the learning rate across $\{0.01, 0.05, 0.001, 0.005, 0.0001, 0.0005\}$. For all datasets and forecasters, the input sequence length was fixed at 96, while prediction lengths were set to $O \in \{96, 192, 336, 720\}$. Comparative results from benchmark forecasting models were derived from K. Yi et al.\cite{14} Datasets were split into training, validation, and test sets at ratios of 6:2:2 for the ETT dataset and 7:1:2 for all other datasets.

\textbf{Baselines:} We selected SOTA forecasters as baselines. Specifically, we incorporated two filter-based models from the FilterNet framework.\cite{14} For Transformer-based approaches, we included two cutting-edge architectures: iTransformer and PatchTST.\cite{11}\cite{12} Additionally, we incorporated TimesNet,\cite{17} a decompositional forecaster that separates time series into trend and residual components to handle complex periodic patterns, and DLinear,\cite{13} a minimalist model that utilizes a simple yet effective MLP-based decoding mechanism. This diverse set of baselines ensures comprehensive coverage of both traditional MLP-based methods and advanced Transformer variants, providing a rigorous comparative benchmark for our study.

\subsection{Main Results}

Table \ref{tab:main} tabulates the results of XLinear compared with baseline forecasters across all datasets. The experimental results demonstrate that our proposed forecaster significantly outperforms state-of-the-art baselines with statistically significant improvements. The detailed analyses are as follows.

\begin{figure}[h]
    \centering
    \includegraphics[width=1\linewidth]{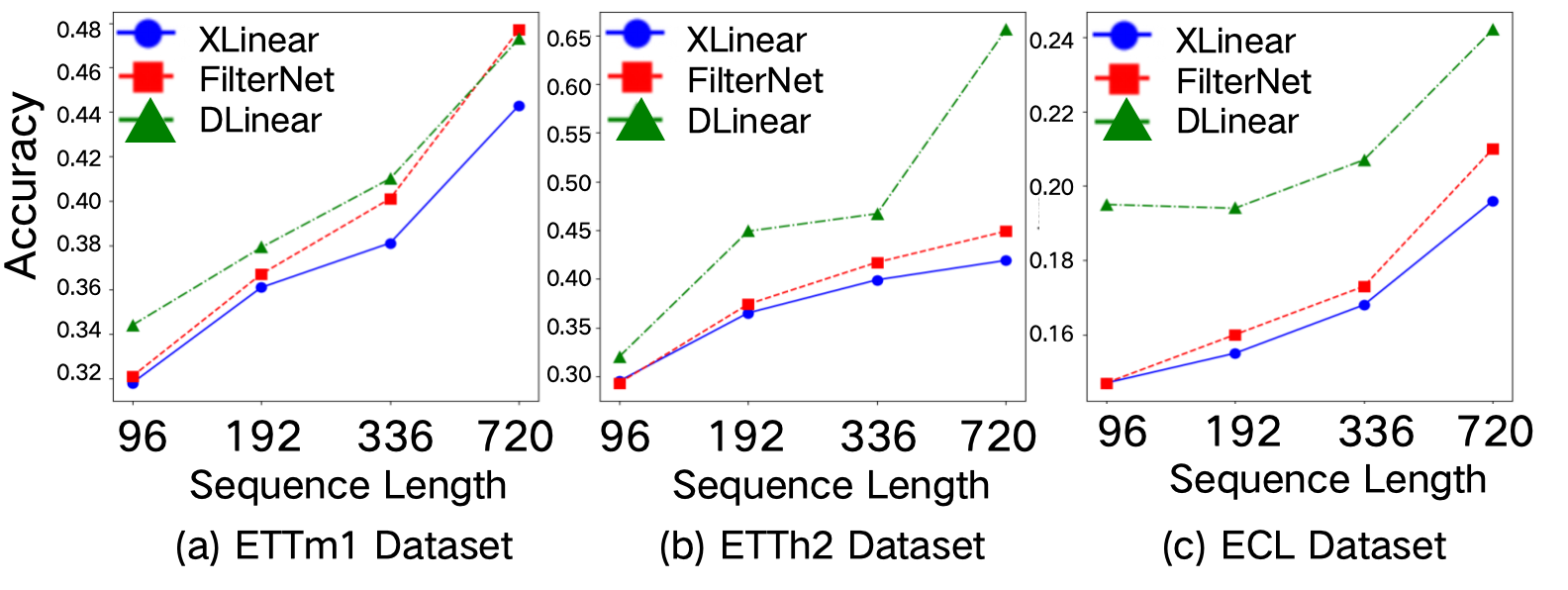}
    \caption{Results of different sequence length on 3 datasets.}
    \label{fig:cmp}
\end{figure}

\textbf{Long-range Forecasting Performance:}
XLinear exhibits remarkable superiority in handling long prediction horizons, especially when the prediction length $L$ is set to 336 or 720, as illustrated in Figure \ref{fig:cmp}. This indicates that our architecture effectively captures long-range dependencies via the EFA block, thereby enhancing its long-range forecasting capability.

\textbf{Robustness Analysis:}
Furthermore, XLinear outperforms baselines across most datasets and all prediction horizons. This highlights that our forecaster maintains robust performance despite integrating the attention mechanism, demonstrating greater versatility across all datasets compared to both baseline methods (TexFilter and PaiFilter).

\textbf{Limitations:}
Notably, while iTransformer outperforms other methods on the Traffic dataset, demonstrating that Transformer-based architectures still hold advantages over MLP-based forecasters in modeling complex datasets, our forecaster has significantly narrowed this performance disparity.

\subsection{Model Analysis}
In this section, we design experiments to prove the validity of the blocks in XLinear and conduct tests on parameter count and inference speed to further validate the efficacy of XLinear.



\begin{figure}[h]
    \centering
    \includegraphics[width=1\linewidth]{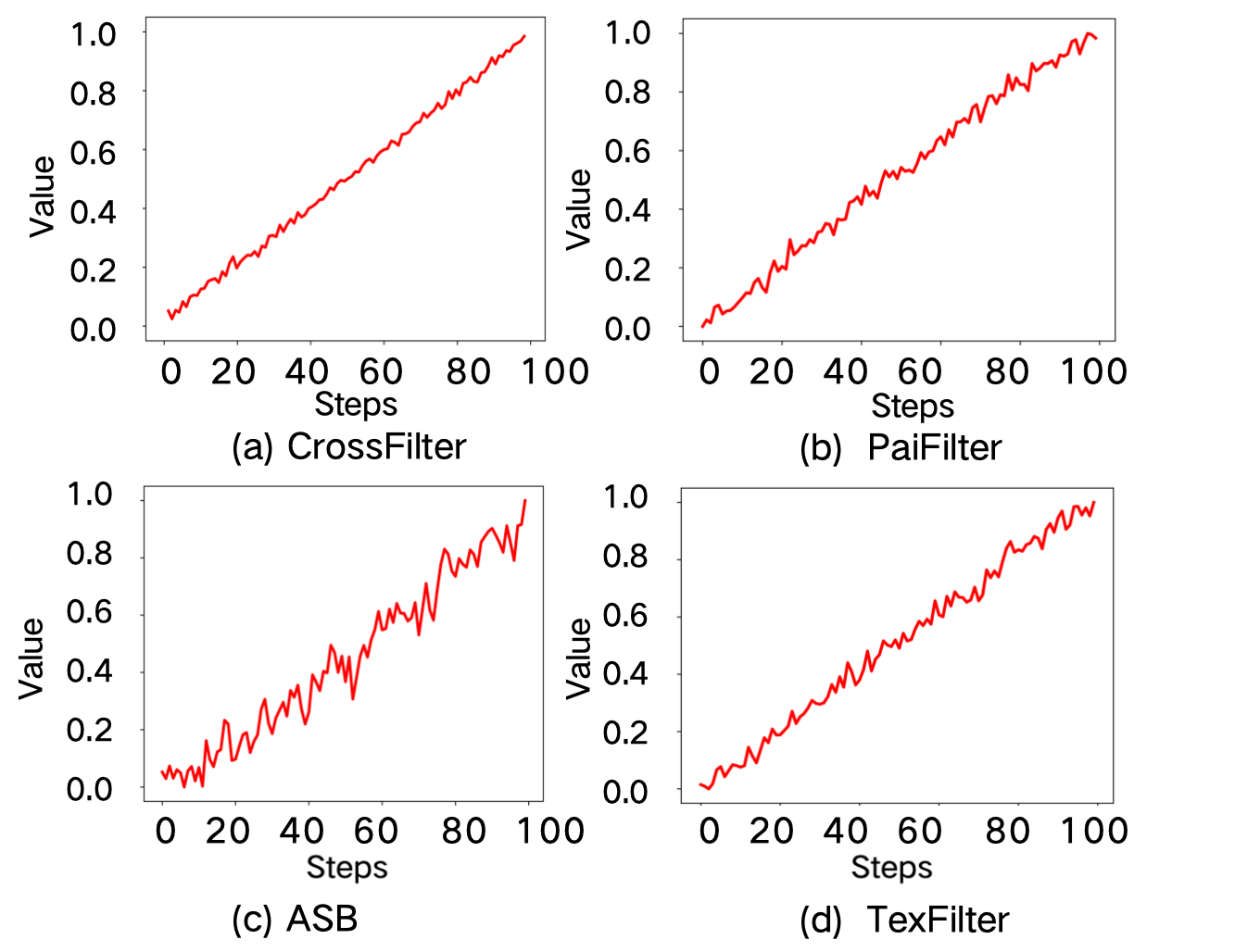}
    \caption{Visualization of filter results. 100 of steps are presented here. ASB refers to Adaptive Spectral Block.\cite{15} A portion of the results was extracted for demonstration.}
    \label{fig:filter}
\end{figure}

\textbf{Modeling Capability of Frequency Filters:}
Our CrossFilter integrates the advantages of PaiFilter and TexFilter, achieving a dual improvement in generalization ability and performance. Comparative experiments demonstrate that this fusion scheme significantly outperforms single filters (PaiFilter and TexFilter). To further verify the denoising capability, we designed a visualization experiment. A hybrid architecture of "filter + LSTM" was employed to model sine wave sequences with added Gaussian noise. The sine function contains 20 periods and is discretized into 10,000 steps. We trained the "filter + LSTM" with input length $L=1000$ and predict length $O = 1000$. Then, we visualized the filtering results of test data in Figure \ref{fig:filter}. The results show that CrossFilter effectively removes noise interference, with denoising performance significantly superior to comparative methods including PaiFilter, TexFilter, and ASB.\cite{15}

\begin{figure}[h]
    \centering
    \includegraphics[width=1\linewidth]{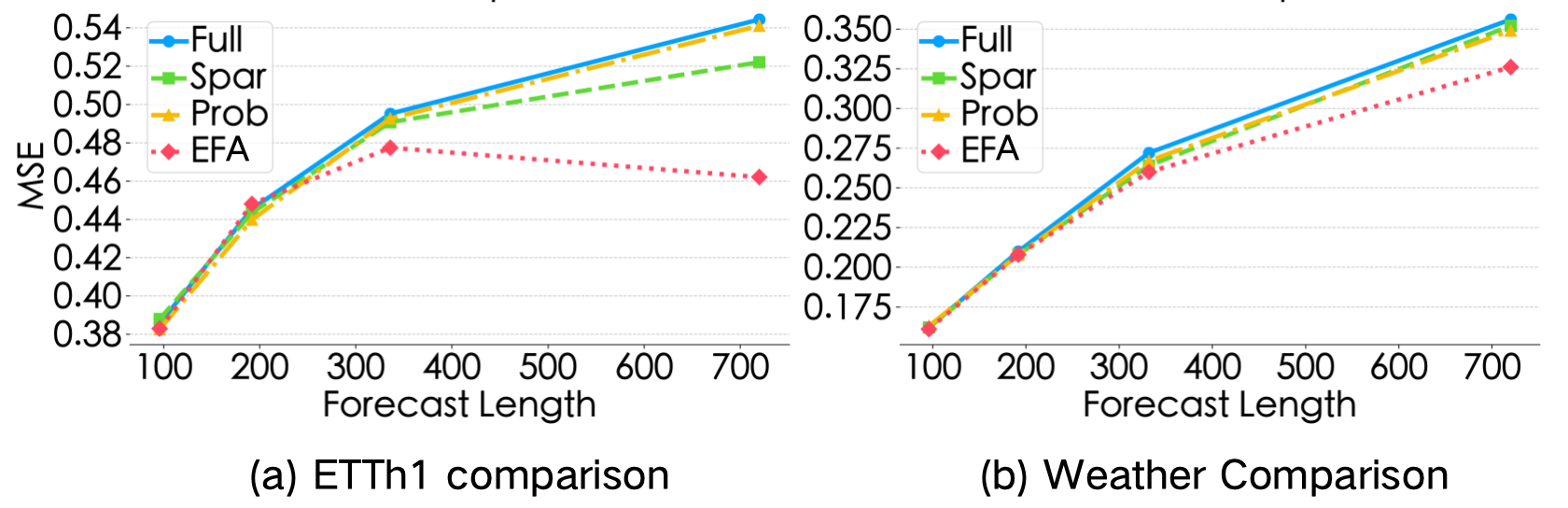}
    \caption{Test on different attention mechanisms. The Full is normal Multi-head Attention, Spar is Sparse Attention, Prob is ProbAttention.\cite{9}\cite{19}\cite{20}}
    \label{fig:attention comp}
\end{figure}

\textbf{Long-range Independence:}
EFA is an attention mechanism designed to capture long-range dependencies and enhance trends in time series data. To verify its effectiveness, we designed comparative experiments where EFA in XLinear was replaced with other attention mechanisms for prediction, and the results are shown in the Figure \ref{fig:attention comp}. The results indicate that although other attention mechanisms exhibit comparable performance to EFA in short-term sequence prediction, their effectiveness is inferior to EFA in long-term sequence prediction. 

\begin{figure}[h]
    \centering
    \includegraphics[width=1\linewidth]{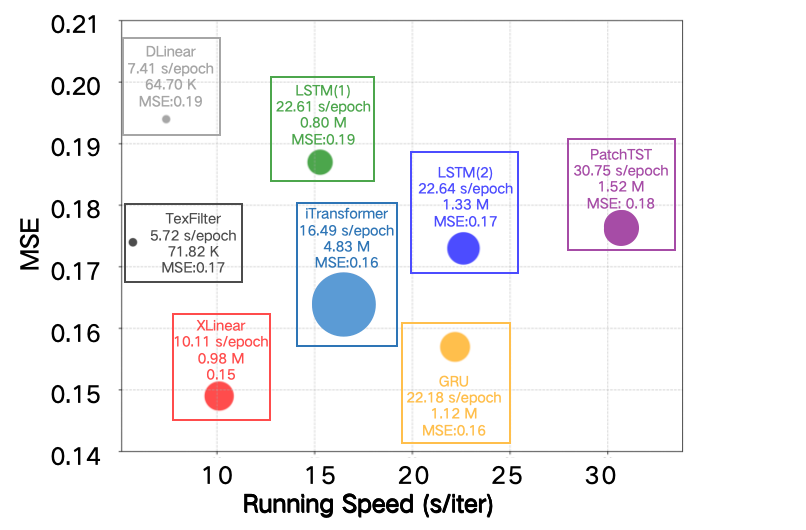}
    \caption{Efficiency comparison.The size of the points indicates the amount of parameters.}
    \label{fig:efficiency}
\end{figure}

\textbf{Efficiency Analysis:}
To rigorously evaluate computational efficiency, we assess XLinear across two dimensions: memory footprint and training throughput. Under identical experimental settings (lookback window of 96 time steps and prediction horizon of 96 steps), we benchmark XLinear against representative baselines, including Transformer-based, MLP-based, and RNN-based forecasters. Results are presented in Figure \ref{fig:efficiency}, where LSTM (1) and LSTM (2) denote single-layer and two-layer LSTM architectures, respectively. Empirical evidence shows that XLinear achieves state-of-the-art accuracy while maintaining a significantly smaller parameter count and faster inference speed compared to Transformer-based models. Although EFA introduces additional parameters relative to vanilla MLP architectures, resulting higher parameter count than simple MLP-based forecaster, XLinear outperforms all MLP-based counterparts in accuracy.

\section{CONCLUSION}
This paper proposed an enhanced MLP-based forecaster namely XLinear to address the limitation of MLP-based forecasters in capturing long-range dependencies. We designed the EFA for frequency enhance and CrossFilter to keep robustness. The experiment results demonstrate that our forecaster outperformed other sota MLP-based forecasters especially in long-range forecasting. In the ablation experiments, we further demonstrate the effectiveness of the XLinear. We test different attention mechanisms and the EFA work best in frequency enhancement. Meanwhile, visualization results of CrossFilter filtering demonstrate that our filter outperforms other filters.

\vspace*{-8pt}
\end{document}